\documentclass[10pt, conference, twocolumn, compsocconf]{IEEEtran}

\usepackage{cite}
\usepackage{amsmath,amssymb,amsfonts}
\usepackage{graphicx}
\usepackage{textcomp}

\usepackage[dvipsnames]{xcolor}

\usepackage{amsthm}
\usepackage[ruled]{algorithm}
\usepackage[noend]{algorithmic}
\usepackage{alltt}
\usepackage{caption}
\usepackage{paralist}
\usepackage{nicefrac}
\usepackage{booktabs}
\usepackage{xspace}
\usepackage{mathtools}
\usepackage{marvosym}
\usepackage{wasysym}
\usepackage{verbatim}
\usepackage{subfigure}
\usepackage{hyperref}
\usepackage{multirow}
\usepackage[capitalize]{cleveref}
\usepackage[per-mode=symbol,detect-all]{siunitx}
\usepackage{graphics}
\usepackage{wrapfig}
\usepackage{enumitem}
\usepackage{float}
\usepackage{courier}
\usepackage{bm}
\usepackage{xfp}
\usepackage{makecell}

\usepackage{stackengine}
\usepackage[normalem]{ulem}
\usepackage{mathpartir}

\newtheorem{lemma}{Lemma}

\theoremstyle{remark}

\usepackage{caption}
\definecolor{color0}{RGB}{228,87,46}
\definecolor{color1}{RGB}{23,190,187}
\definecolor{color2}{RGB}{255,201,20}
\definecolor{color3}{RGB}{46,40,42}
\definecolor{color4}{RGB}{118,176,65}

\definecolor{color0}{RGB}{250,121,33}
\definecolor{color1}{RGB}{254,153,32}
\definecolor{color2}{RGB}{185,164,76}
\definecolor{color3}{RGB}{86,110,61}
\definecolor{color4}{RGB}{12,71,103}

\definecolor{color0}{RGB}{228,253,225}
\definecolor{color1}{RGB}{138,203,136}
\definecolor{color2}{RGB}{100,131,129}
\definecolor{color3}{RGB}{87,87,97}
\definecolor{color4}{RGB}{255,191,70}

\definecolor{color0}{RGB}{226,59,62}
\definecolor{color1}{RGB}{243,114,44}
\definecolor{color2}{RGB}{248,150,30}
\definecolor{color3}{RGB}{249,199,79}
\definecolor{color4}{RGB}{126,179,86}
\definecolor{color5}{RGB}{67,170,139}
\definecolor{color6}{RGB}{39,125,161}
\definecolor{color7}{RGB}{21,49,60}
\definecolor{color8}{RGB}{180,215,228}

\definecolor{nnedgecolor}{RGB}{90,90,90}

\usepackage{tikz,ifthen,pgfplots}
\usetikzlibrary{arrows,trees,backgrounds,automata,shapes,decorations,plotmarks,fit,calc,positioning,shadows,chains}
\usetikzlibrary{arrows.meta,bending}
\tikzstyle{every pin edge}=[<-,shorten <=1pt]
\tikzstyle{every path}=[draw=color7!50]
\tikzstyle{neuron}=[circle,fill=black!25,minimum size=17pt,inner sep=0pt]
\tikzstyle{input neuron}=[neuron, fill=color4]
\tikzstyle{output neuron}=[neuron, fill=color0]
\tikzstyle{hidden neuron}=[neuron, fill=color6!80]
\tikzstyle{annot} = [text width=4em, text centered]
\tikzstyle{nnedge} = [-{stealth},shorten >=0.1cm, shorten <=0.05cm,line width=0.8pt,nnedgecolor]
\usetikzlibrary{calc}

\newcommand{\relu}{\text{ReLU}\xspace}
\newcommand{\argmax}{\text{argmax}\xspace}

\newcommand{\kfr}{$k$-forward-redundant}
\newcommand{\kfracy}{$k$-forward-redundancy}

\newcommand{\errorMinimizerFunction}{l_m}

\newcommand{\sat}{\texttt{SAT}}
\newcommand{\unsat}{\texttt{UNSAT}}

\usepackage{bussproofs}

\newif\ifcomments
\commentstrue

\newif\ifoutline
\outlinefalse

\newif\iflong
\longtrue

\renewcommand{\paragraph}[1]{\vspace{1mm}\noindent{\bf #1}\ }

\newcommand\blfootnote[1]{%
  \begingroup
  \renewcommand\thefootnote{}\footnote{#1}%
  \addtocounter{footnote}{-1}%
  \endgroup
}

\begin{document}

\title{Pruning and Slicing Neural Networks\\
using Formal Verification}

\author{
\IEEEauthorblockN{Ori Lahav and Guy Katz}
\IEEEauthorblockA{The Hebrew University of Jerusalem, Jerusalem, Israel}
\{ori.lahav, guykatz\}@cs.huji.ac.il
}

\maketitle

\begin{abstract}
\blfootnote{[*] This is the extended version
of a paper with the same title that is about to appear in FMCAD 2021.
See \url{https://fmcad.org/}}
  Deep neural networks (DNNs) play an increasingly important role in
  various computer systems. In order to create these networks,
  engineers typically specify a desired topology, and then use an
  automated training algorithm to select the network's weights. While
  training algorithms have been studied extensively and are well
  understood, the selection of topology remains a form of art, and can
  often result in networks that are unnecessarily large --- and
  consequently are incompatible with end devices that have limited
  memory, battery or computational power. Here, we propose to address
  this challenge by harnessing recent advances in DNN verification.
  We present a framework and a methodology for discovering
  redundancies in DNNs --- i.e., for finding neurons that are not
  needed, and can be removed in order to reduce the size of the DNN.
  By using sound verification techniques, we can formally guarantee
  that our simplified network is equivalent to the original, either
  completely, or up to a prescribed tolerance. Further, we show how to
  combine our technique with \emph{slicing}, which results in a
  \emph{family} of very small DNNs, which are together equivalent to
  the original. Our approach can produce DNNs that are significantly
  smaller than the original, rendering them suitable for deployment on
  additional kinds of systems, and even more amenable to subsequent
  formal verification.  We provide a proof-of-concept implementation
  of our approach, and use it to evaluate our techniques on several
  real-world DNNs.

\end{abstract}

\section{Introduction}

The wide-spread adoption of \emph{deep learning}~\cite{GoBeCo16} has
caused a significant leap forward in many domains within computer
science.  \emph{Deep neural networks} (\emph{DNNs}) have now become
the state of the art solution for a myriad of real-world problems,
such as game playing~\cite{SiHuMaGuSiVaScAnPaLaDi16}, image
recognition~\cite{SiZi14}, and autonomous
vehicles~\cite{BoDeDwFiFlGoJaMoMuZhZhZhZi16,JuLoBrOwKo16}. This trend
is likely to continue and intensify, thus creating an urgent need for
tools and techniques to analyze and manipulate DNNs.

A part of the appeal of DNNs is that they are produced in a mostly
automated way.  In order to create a DNN for a particular task at
hand, engineers first specify the network architecture ---
specifically, the number of layers in the network, the size and type
of each layer, and the inter-layer connections. Then, they invoke an
automated training algorithm for assigning weights to the network's
edges\cite{GoBeCo16}. While the automated training process has been
extensively studied and is generally well understood~\cite{GoBeCo16},
the choice of network architecture is still performed according to
various rules of thumb, and is considered a form of art. This can
often lead to a choice of architecture that is wasteful --- i.e.,
which results in a large DNN, whereas a smaller DNN could have
achieved similar
accuracy~\cite{GoFeMaBaKa20,HaMaDa15,IaHaMoAsDaKe16}. For DNNs
intended to run on devices with limited resources (e.g., mobile
phones, or embedded circuits), excessive DNN size can be a limiting
factor~\cite{JuLoBrOwKo16}.

One successful approach for mitigating this difficulty is to first
train a large network, and then shrink it by removing \emph{redundant
  neurons}.  Informally, we say that a neuron is redundant if removing
it does not change the DNN's output; and thus, removing
it from a network $N$ results in a smaller network, $N'$, that is
\emph{equivalent} to $N$.  In order to identify redundant neurons within a
DNN, prior work has focused primarily on \emph{heuristic pruning}:
heuristically identifying neurons and edges that contribute little to
the network's output, removing these neurons, and then performing
additional training of the
network~\cite{HaMaDa15,IaHaMoAsDaKe16}. These methods have been highly
successful in reducing DNN sizes, but they provide no formal
guarantees; i.e., the removed neurons are not guaranteed to have been
redundant, and the simplified network can thus be dramatically
different from the original, producing different results for various
inputs~\cite{liebenwein2021lost}.

Recently, there has been a surge of interest in the formal
verification of neural networks
(e.g.,~\cite{KaBaDiJuKo17,KaBaDiJuKo21,KuKaGoJuBaKo18,HuKwWaWu17,GeMiDrTsChVe18,WaPeWhYaJa18,AmWuBaKa21}, and
many others). These new capabilities have made it possible to identify
and remove redundancies in a network, in a way that \emph{guarantees}
that the smaller network is completely equivalent to the
original~\cite{GoFeMaBaKa20}. Specifically, Gokulanathan et al.~showed
how verification could be used to identify and remove ``dead''
neurons, i.e. neurons whose output is $0$ regardless of the network's
inputs. This approach was shown to reduce network sizes by up to
$10\%$, which is quite significant, while preserving complete
equivalence to the original network.


Here, we propose a new technique, which also attempts to apply formal
verification in order to remove neurons from a DNN, but which is
significantly stronger. Specifically, our technique:
\begin{inparaenum}[(i)]
\item can identify additional kinds of redundant neurons (beyond
  ``dead'' neurons), whose removal does not affect the network's
  outputs at all; and
\item can identify additional redundant neurons, whose removal
  \emph{does} affect the network's outputs, but only up to a small,
  provable bound.
\end{inparaenum}





Finally, we propose a method that takes our approach to the extreme,
by integrating it with \emph{network slicing}. This method, in which a
network is simplified into a family of much smaller sub-networks, is
appropriate for cases where fast inference is crucial: an input is
checked to identify the appropriate sub-network for handling it, and
then only that network needs to be evaluated for that specific input.
Slicing is achieved by partitioning the DNN's input
domain into small sub-domains, maintaining a separate DNN for each
input sub-domain, and then applying the aforementioned simplification
techniques on each of these DNNs. We demonstrate that the use of 
small input sub-domains causes many neurons to become redundant, and
consequently removable.

For evaluation purposes, we implemented our approach in an
open-source, publicly available tool~\cite{ourCode}. As a
backend, our tool uses the Marabou DNN verification
tool~\cite{Marabou2019}. We note, however, that our approach is
agnostic of the underlying verification engine --- indeed, it could be
integrated with any other tool, and will consequently benefit from any
development in DNN verification technology.  We evaluated our approach
on a set of airborne collision avoidance networks~\cite{JuLoBrOwKo16},
obtaining highly favorable results. Specifically, we were able to
achieve a reduction of up to $71$\% in overall network sizes, while
keeping the outputs identical (up to a prescribed tolerance) to those
produced by the original DNN.  This reduction in network sizes is a
significant improvement over the previous state of the
art~\cite{GoFeMaBaKa20}. Further, while prior techniques were
specifically tailored to networks with only a specific activation
function (i.e., rectified linear units~\cite{GoFeMaBaKa20}), our
technique is applicable to multiple kinds of DNNs.

The rest of this paper is organized as follows. In
Section~\ref{sec:background}, we provide the necessary background on
DNNs and their verification. Next, in Section~\ref{sec:formalization}
we present the basic building block of our approach, namely the
removal of a single neuron. We then specify multiple kinds of neurons
that can be removed in Section~\ref{sec:linear_funcs}, and discuss the
simultaneous removal of neurons in
Section~\ref{sec:simultaneousRemoval}.  Subsequently, in
Section~\ref{sec:input-slicing} we present how \emph{input slicing}
and simplification can be used to improve network evaluation time.  An
evaluation appears in Section~\ref{sec:evaluation}, followed by a
discussion of related work in Section~\ref{sec:relatedWork}. We then
conclude in Section~\ref{sec:conclusion}.

\section{Background: DNNs and their Verification}
\label{sec:background}

A deep neural network~\cite{GoBeCo16} is a directed, acyclic graph,
whose nodes (also referred to as \emph{neurons}) are grouped into
layers. The first layer is the \emph{input layer}; the final layer is
the \emph{output layer}; and the intermediate layers are the
\emph{hidden layers}. When the network is evaluated, the input neurons
are assigned some values (e.g., sensor readings), and these values are
then propagated through the network, layer by layer, until the output
values are computed. In \emph{regression} networks, the numeric value
of the output is of interest, while in the case of
\emph{classification} networks, the output
neurons correspond to possible \emph{labels} that the network can
classify the input into; and the label whose neuron obtained the
highest score is the one returned by the network.

\begin{figure}[htp]
\centering %
\scalebox{0.65}{%
\noindent\begin{tikzpicture}
\def\layersepedges{2cm}
\def\layersep{4cm}
\def\forwardsep{1.8cm}
\def\vertSepFactoryI{0.7}
\def\vertSepFactory{2}
\def\shiftFactory{1.9cm}
    \foreach \name / \y in {1,...,5}
        \node[input neuron, label={[label distance=4mm]90:\ifthenelse{\y=1}{Inputs}{}}] (I-\name) at (0,-\vertSepFactoryI * \y) {};

    \foreach \name / \y in {1,...,3}
        \path[yshift=\shiftFactory]
            node[hidden neuron, label={[label distance=2mm]\ifthenelse{\y=1}{\bf{WS}}{}}] (B1-\name) at (\layersepedges + 0*\layersep,-\vertSepFactory * \y cm) {};
            
    \foreach \name / \y in {1,...,3}
        \path[yshift=\shiftFactory]
            node[hidden neuron, label={[label distance=2mm]\ifthenelse{\y=1}{\bf{ReLU}}{}}] (F1-\name) at (\layersepedges + 0*\layersep + \forwardsep,-\vertSepFactory * \y cm) {};

    \foreach \name / \y in {1,...,3}
        \path[yshift=\shiftFactory]
            node[hidden neuron, label={[label distance=2mm]\ifthenelse{\y=1}{\bf{WS}}{}}] (B2-\name) at (\layersepedges + 1*\layersep,-\vertSepFactory * \y cm) {};

    \foreach \name / \y in {1,...,3}
        \path[yshift=\shiftFactory]
            node[hidden neuron, label={[label distance=2mm]\ifthenelse{\y=1}{\bf{ReLU}}{}}] (F2-\name) at (\layersepedges + 1*\layersep + \forwardsep,-\vertSepFactory * \y cm) {};

    \foreach \name / \y in {1,...,3}
        \path[yshift=\shiftFactory]
            node[hidden neuron, label={[label distance=2mm]\ifthenelse{\y=1}{\bf{ReLU}}{}}] (F3-\name) at (\layersepedges + 1*\layersep + 2cm + \forwardsep,-\vertSepFactory * \y cm) {};

    \foreach \name / \y in {1,...,5}
        \node[output neuron, label={[label distance=4mm]90:\ifthenelse{\y=1}{Outputs}{}}]
        (O-\name) at (\layersepedges*2 + 1*\layersep + 2cm + \forwardsep,-\vertSepFactoryI * \y cm) {};

    \foreach \source in {1,...,5}
        \foreach \dest in {1,...,3}
            \draw[nnedge] (I-\source.east) -- (B1-\dest.west);
            
    \foreach \source in {1,...,3}
        \foreach \dest in {1,...,5}
            \draw[nnedge] (F3-\source.east) -- (O-\dest.west);

    \foreach \source in {1,...,3}
        \foreach \dest in {1,...,3}
            \draw[nnedge] (F1-\source) -- (B2-\dest.west);

    \foreach \source in {1,...,3}
        \draw[nnedge] (B1-\source) -- node[above]{ReLU} (F1-\source);
    	
    \foreach \source in {1,...,3}
        \draw[nnedge] (B2-\source) -- node[above]{ReLU} (F2-\source);
        
	\node [xshift=\forwardsep/2,yshift=0.25cm,rotate=90] at ($(B1-1)!.5!(B1-2)$) {\textbf{\ldots}};
	\node [xshift=\forwardsep/2,yshift=0.25cm,rotate=90] at ($(B1-2)!.5!(B1-3)$) {\textbf{\ldots}};
	
	\node [xshift=\forwardsep/2,yshift=0.25cm,rotate=90] at ($(B2-1)!.5!(B2-2)$) {\textbf{\ldots}};
	\node [xshift=\forwardsep/2,yshift=0.25cm,rotate=90] at ($(B2-2)!.5!(B2-3)$) {\textbf{\ldots}};
	
	\node [rotate=90] at ($(F3-1)!.5!(F3-2)$) {\textbf{\ldots}};
	\node [rotate=90] at ($(F3-2)!.5!(F3-3)$) {\textbf{\ldots}};
	\node at ($(F2-2)!.5!(F3-2)$) {\textbf{\ldots}};

\end{tikzpicture}
}
\caption {An illustration of a DNN with alternating weighted-sum (WS) and ReLU layers.}
\label{fig:dnn}
\end{figure}
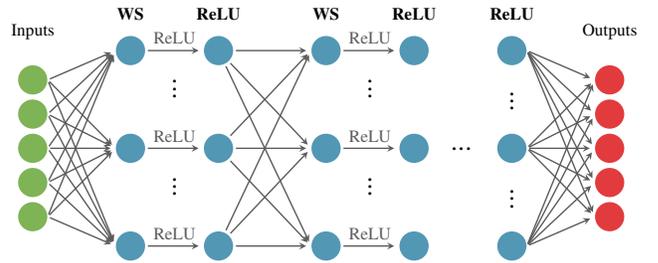

Each layer in the DNN has a type, which determines how its neuron
values are computed. Here, we will focus on two types: \emph{weighted
  sum} layers, and \emph{piecewise-linear activation} layers.  In a
weighted-sum layer, the value of a neuron $y$ is computed as
$y=b + \sum c_iv_i$ for neurons $v_i$ from preceding layers, where the
\emph{weights} $c_i$ are determined when the network is first
trained. In a piecewise-linear activation layer, the value of neuron
$y$ is computed as
\begin{equation*}
y=
\begin{cases}
  a_1x+b_1 & \text{if } s_1\leq x < s_2,\\
  a_2x+b_2 & \text{if } s_2\leq x < s_3,\\
  \ldots \\
  a_kx+b_k & \text{if } s_k\leq x \leq s_{k+1}
\end{cases}
\end{equation*}
where $x$ is a neuron from some preceding layer, and the $a_i$, $b_i$ and
$s_i$ parameters determine the piecewise linear function being computed. A
common example of a piecewise-linear activation function is the
\relu{} function, given by
\begin{wrapfigure}[9]{r}{3.8cm}
  \vspace{-0.5cm}
  \begin{center} 
\includegraphics[scale=0.25]{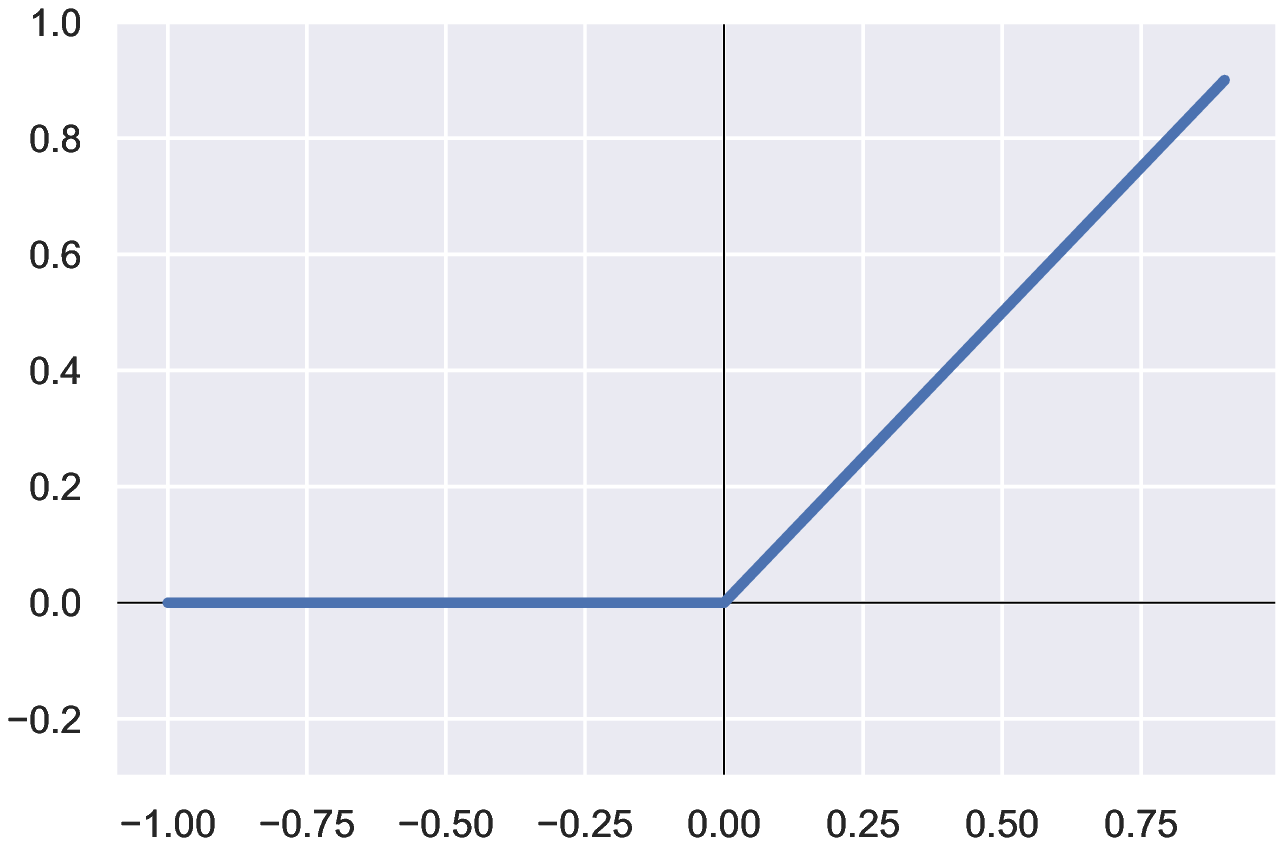}
    \captionsetup{size=small}
    \captionof{figure}{The \relu{} function.}
    \label{fig:relu}
  \end{center}
\end{wrapfigure}
\vspace{-0.3cm}
\begin{equation*}
y=\max(x,0)=
\begin{cases}
  0 & \text{if } x < 0 \\
  x & \text{if } x \geq 0 
\end{cases}
\end{equation*}
(see Fig.~\ref{fig:relu}). Together, weighted-sum layers and
piecewise-linear activation functions make up many common DNN
architectures~\cite{GoBeCo16}. Typically, they are used in alternation
(see Fig.~\ref{fig:dnn}). Extending our approach to activation
functions that are not piecewise-linear remains a work in progress.


More formally, we regard a DNN $N$ with $k$ inputs and $m$ outputs as
a mapping $\mathbb{R}^k\rightarrow\mathbb{R}^m$. The DNN is given as a
sequence of layers $L_1,\ldots, L_n$, where $L_1$ is the input layer
and $L_n$ is the output layer. We use $s_i$ to denote the size of
layer $L_i$, and use $v_i^1,\ldots,v_i^{s_i}$ to refer to the
individual neurons of $L_i$. We use $V_i$ to refer to the column
vector $[v_i^1,\ldots,v_i^{s_i}]^T$.  When the network is being
evaluated, we assume that the input values $V_1$ are given, and that
$V_2,\ldots,V_n$ are computed iteratively. The type of each hidden
layer is given via the mapping
$T_N:\mathbb{N}\rightarrow\mathcal{T}$. For simplicity we set
$\mathcal{T}=\{\text{weighted-sum}, \relu{}\}$, although our technique
applies to all types of piecewise-linear activation functions.

In a weighted-sum layer $L_i$, each neuron $v_i^j$ is associated with
a linear function $v_i^j=b_i^j + \sum c_{l,t}\cdot v_l^t$; i.e.,
$v_i^j$ is computed as a weighted-sum of neurons $v_l^t$ from
preceding layers $l<i$, plus a bias value $b_i^j$. In a ReLU layer
$L_i$, each neuron $v_i^j$ is associated with a specific neuron
$v_l^t$ from a preceding layer $l<i$, and its value is given by
$v_i^j=\relu{}(v_l^t)=\max(v_l^t, 0)$. Note that each neuron's value
depends only on neurons from preceding layers.

In recent years, various security and safety issues have been
discovered in DNNs~\cite{SzZaSuBrErGoFe13,KaBaDiJuKo17}. This has led
the verification community to study the \emph{DNN verification
  problem}~\cite{LiArLaBaKo19}. Generally, this problem is defined by
a set of constraints $P$ on the DNN's inputs, and a set of constraints
$Q$ on the DNN's outputs; and solving it entails finding (or proving
the non-existence of) an input $x$ such that $P(x)\wedge Q(N(x))$;
i.e., an input $x$ that satisfies the input condition, and is mapped
by the DNN to a point that satisfies the output condition. When $P$
and $Q$ characterize an unsafe behavior of the DNN, an $\unsat{}$
answer to the aforementioned query indicates that the DNN is safe;
whereas a $\sat{}$ answer, accompanied by a satisfying assignment,
demonstrates an unsafe behavior. This formalization is sufficiently
expressive for capturing many properties of
interest~\cite{KaBaDiJuKo17}.  Many approaches for solving the DNN
verification problem have been proposed recently
(e.g.,~\cite{KaBaDiJuKo17,HuKwWaWu17,GeMiDrTsChVe18,WaPeWhYaJa18}, and
many others). The techniques we discuss in this work use a DNN
verification engine as a backend, and do not depend on the precise
method used --- and so we do not elaborate on this topic. We refer the
interested reader to~\cite{LiArLaBaKo19} for a survey.

\section{Removing a Single Neuron}
\label{sec:formalization}

The core of our DNN simplification approach is the identification, and
then the removal, of \emph{redundant neurons}. Given a DNN $N$, we seek to
identify a redundant neuron $v_i^j$, and then produce another network,
$N'$, which is identical to $N$ except for the redundant neuron that
has been removed. Ideally, we would like to ensure that $N$ and $N'$ are
equivalent; i.e., that $\forall x. N(x)=N'(x)$. Because $N'$ is
obtained from $N$ by removing a neuron, it is smaller; and this
process can be repeated iteratively, to eventually obtain a
significantly smaller network that is equivalent to $N$.
Of course, the key points that need addressing are:
\begin{inparaenum}[(i)]
\item how to technically remove a redundant neuron from the network; and
\item how to identify redundant neurons.
\end{inparaenum}
In this section we focus on the first challenge, and describe the
mechanics of removing a neuron.

In order to maintain compatibility with the original network, we will
refrain from removing neurons from the network's input or output
layers; all other neurons are considered candidates for removal. We
distinguish between neurons in weighted-sum layers, and neurons in
activation function layers. In fact, our proposed approach only
supports the removal of weighted-sum neurons that feed only into other
weighted-sum neurons; and the removal of activation function neurons
will be performed by first transforming them into weighted-sum
neurons, as described in later sections.

Consider a neuron $v$ computed as a weighted-sum
\[
  v= b_v + \sum c_i \cdot x_i,
\]
where $x_i$ are neurons from preceding layers.
 Suppose that $v$ only feeds into other weighted-sum neurons, and let
 $u$ be such a neuron:
\[
  u = b_u  + c\cdot v +  \sum d_i \cdot y_i,
\]
where $y_i$ are again neurons from preceding layers. 
In this case, 
$u$'s equation can be updated into:
\[
  u = (b_u + c\cdot b_v) + \sum c\cdot c_i\cdot x_i + \sum d_i\cdot y_i.
\]
If this process is repeated for every (weighted-sum) neuron that $v$
feeds into, then afterwards $v$ will have no outgoing
edges. Consequently, $v$ could then  be eliminated from the network altogether.
It is straightforward to show that such an operation will never affect
the value of $u$, and that the modified network will thus be
completely equivalent to the original. Also, identifying neurons that
can be eliminated is simple, and amounts to searching for weighted-sum
neurons that are only connected to other weighted-sum neurons.

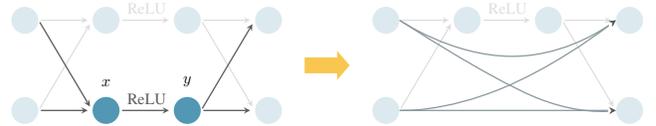
\begin{figure}[htp]
\centering%
\scalebox{0.6}{%
\begin{tikzpicture}
\def\layersepedges{2cm}
\def\layersep{4cm}
\def\forwardsep{1.8cm}
\def\vertSepFactoryI{0.7}
\def\vertSepFactory{2}
\def\shiftFactory{1.9cm}
\def\opac{0.2}
\def\shift{8cm}

    \foreach \name / \y in {1,...,2}
        \path[opacity=\opac, yshift=\shiftFactory]
            node[hidden neuron] (F0-\name) at (0,-\vertSepFactory * \y cm) {};
            
    \path[opacity=\opac, yshift=\shiftFactory]
        node[hidden neuron] (B1-1) at (\forwardsep,-\vertSepFactory * 1 cm) {};
    \path[yshift=\shiftFactory]
        node[hidden neuron, label={[label distance=1mm]90:$x$}] (B1-2) at (\forwardsep,-\vertSepFactory * 2 cm) {};

    \path[opacity=\opac, yshift=\shiftFactory]
        node[hidden neuron] (F1-1) at (\forwardsep*2,-\vertSepFactory * 1 cm) {};
    \path[yshift=\shiftFactory]
        node[hidden neuron, label={[label distance=1mm]90:$y$}] (F1-2) at (\forwardsep*2,-\vertSepFactory * 2 cm) {};

    \foreach \name / \y in {1,...,2}
		\path [opacity=\opac, yshift=\shiftFactory]
            node[hidden neuron] (B2-\name) at (\forwardsep*3,-\vertSepFactory * \y cm) {};

    \foreach \source in {1,...,2} {
        \draw [opacity=\opac] (F0-\source.east) edge [nnedge] (B1-1.west);
        \draw (F0-\source.east) edge [nnedge] (B1-2.west);
    }
    
    \path[opacity=\opac] (B1-1.east) edge[nnedge] node[above]{ReLU} (F1-1);
    \path (B1-2.east) edge[nnedge] node[above]{ReLU} (F1-2);

    \foreach \dest in {1,...,2} {
        \path[opacity=\opac] (F1-1.east) edge[nnedge] (B2-\dest.west);
        \path (F1-2.east) edge[nnedge] (B2-\dest.west);
    }

	\draw[-{Triangle[width=18pt,length=8pt]}, line width=10pt, color=color3](6.2,-1.1) -- (7.2, -1.1);

    \foreach \name / \y in {1,...,2}
        \path[opacity=\opac, yshift=\shiftFactory]
            node[hidden neuron] (Ft0-\name) at (\shift,-\vertSepFactory * \y cm) {};
            
    \path[opacity=\opac, yshift=\shiftFactory]
        node[hidden neuron] (Bt1-1) at (\shift+\forwardsep,-\vertSepFactory * 1 cm) {};

    \path[opacity=\opac, yshift=\shiftFactory]
        node[hidden neuron] (Ft1-1) at (\shift+\forwardsep*2,-\vertSepFactory * 1 cm) {};

    \foreach \name / \y in {1,...,2}
        \path[opacity=\opac, yshift=\shiftFactory]
            node[hidden neuron] (Bt2-\name) at (\shift+\forwardsep*3,-\vertSepFactory * \y cm) {};

    \foreach \source in {1,...,2} {
        \path[opacity=\opac] (Ft0-\source.east) edge[nnedge] (Bt1-1.west);
    }
    
    \path[opacity=\opac] (Bt1-1.east) edge[nnedge] node[above]{ReLU} (Ft1-1);

    \foreach \dest in {1,...,2} {
        \path[opacity=\opac] (Ft1-1.east) edge[nnedge] (Bt2-\dest.west);
    }
    
    \draw [-{stealth},line width=0.8pt,nnedgecolor] [thick,draw=color7!50] plot [smooth,tension=1] coordinates {(Ft0-2.east)  ([shift={(-2.5, -1.5)}]Bt2-1.west) (Bt2-1.west)};
    \draw[-{stealth},line width=0.8pt,nnedgecolor] [thick,draw=color7!50][thick,draw=color7!50] plot [smooth,tension=0.6] coordinates {(Ft0-2.east)  (Bt2-2.west)};
    
    \draw[-{stealth},line width=0.8pt,nnedgecolor] [thick,draw=color7!50][thick,draw=color7!50] plot [smooth,tension=1] coordinates {(Ft0-1.east)  ([shift={(-2.3, -0.8)}]Bt2-1.west) (Bt2-1.west)};
    \draw[-{stealth},line width=0.8pt,nnedgecolor] [thick,draw=color7!50][thick,draw=color7!50] plot [smooth,tension=0.8] coordinates {(Ft0-1.east)  ([shift={(-1.8, 0.3)}]Bt2-2.west) (Bt2-2.west)};
\end{tikzpicture}
}
\caption {Illustration: removing a neuron. $x$ is a weighted-sum neuron which feeds into $y$,
a \relu{} neuron. After converting $y$ into a weighted-sum neuron, both $x$ and $y$ can be
removed.\label{fig:removing}}
\end{figure}

In practice, DNN topology usually alternates between weighted-sum and
activation function layers, and so consecutive weighted-sum neurons
are likely to be scarce. Our strategy will thus be to replace
activation function neurons with weighted-sum neurons, in a way that
will enable neuron removal while preserving network accuracy.  As an
example, let us consider a ReLU neuron, $y=\relu{}(x)$.
Because of layer-type alternation, it is reasonable to assume that $x$
is a weighted-sum neuron. In this case, if we can express $y$ as a linear
function of $x$, i.e. $y=ax+b$ for some $a$ and $b$, then the previous
case of two consecutive weighted-sum neurons applies: we can remove
$x$ entirely, change $y$'s type to weighted-sum, and connect $y$ to
$x$'s inputs. Further, if $y$ also feeds into
weighted-sum neurons, then we can apply simplification once again, and
remove $y$ as well. An illustration appears in Fig.~\ref{fig:removing}.

The aforementioned steps constitute the framework of our approach --- to
repeat, until saturation, the two steps:
\begin{inparaenum}[(i)]
\item identify any weighted-sum neurons that only feed into weighted
  sum neurons, and remove them; and
\item identify any activation function neurons that can be changed
  into weighted-sum neurons, without harming the network's accuracy.
\end{inparaenum}
The key remaining issue is how to identify those neurons to which step
2 can be applied. We elaborate on this issue in the following sections.


\section{Linearizing Activation Functions}
\label{sec:linear_funcs}
We next propose various criteria for determining which activation
function neuron can be changed into weighted-sum neurons. Applying
these criteria in practice is discussed later, in
Section~\ref{sec:simultaneousRemoval}.

\medskip
\noindent
\textbf{Phase Redundancy.}  In order to transform an activation
function neuron into a weighted-sum neuron
without changing the network's outputs, we leverage the properties of
piecewise-linear functions. Let $x$ be a weighted-sum neuron
and let $y=f(x)$ be an activation function neuron; then, by
definition, the value range of $x$ is divided into segments
$[s_1,s_2], [s_2,s_3], \ldots [s_k,s_{k+1}]$, and in each segment $y$
is a linear function (a weighted-sum) of $x$.  If we are able to
discover that $x$ is in fact restricted to one of these segments, i.e.
$s_i\leq x< s_{i+1}$ for some $i$, then we can safely discard the
constraint $y=f(x)$ and replace it with a linear constraint
$y=a_ix+b_i$, thus changing $y$ to be a weighted-sum neuron. We stress
that this change does not alter the value of $y$, and consequently
does not alter the network's outputs. When this phenomenon occurs, we
say that $y$ is \emph{phase-redundant}.  For the ReLU function, this
happens if we discover that $x< 0$ ($y$ is
\emph{inactive-redundant}), or $x\geq 0$ ($y$ is
\emph{active-redundant}). As previously stated, transforming the
piecewise-linear constraint into a linear one will often allow us to
eliminate two neurons from the network, without changing its outputs.

\begin{figure}[htp]
\centering %
\scalebox{0.65}{%
\noindent\begin{tikzpicture}
\def\layersepedges{2cm}
\def\layersep{3.3cm}
\def\forwardsep{1cm}
\def\vertSepFactoryI{0.3}
\def\vertSepFactory{1.5}
\def\shiftFactory{1.9cm}

    \node[input neuron, label={[label distance=2mm]90:Input}, label={[label distance=1.5mm]270:$[-1,1]$}] (I-1) at (0,-\vertSepFactoryI * 1) {};

	\path[yshift=\shiftFactory]
            node[hidden neuron, text=white, label={[label distance=2mm]\bf{WS}}] (B1-1) at (\layersepedges + 0*\layersep,-\vertSepFactory * 1 cm) {$+0$};
	\path[yshift=\shiftFactory]
            node[hidden neuron, text=white] (B1-2) at (\layersepedges + 0*\layersep,-\vertSepFactory * 2 cm) {$+1$};

    \path[yshift=\shiftFactory]
            node[hidden neuron, color=color2, text=white, label={[label distance=2mm]\bf{ReLU}}] (F1-1) at (\layersepedges + 0*\layersep + \forwardsep,-\vertSepFactory * 1 cm) {\large $y$};
\path[yshift=\shiftFactory]
            node[hidden neuron] (F1-2) at (\layersepedges + 0*\layersep + \forwardsep,-\vertSepFactory * 2 cm) {};

    \foreach \name / \y in {1,...,2}
        \path[yshift=\shiftFactory]
            node[hidden neuron, text=white, label={[label distance=2mm]\ifthenelse{\y=1}{\bf{WS}}{}}] (B2-\name) at (\layersepedges + 1*\layersep,-\vertSepFactory * \y cm) {$+1$};

    \foreach \name / \y in {1,...,2}
        \path[yshift=\shiftFactory]
            node[hidden neuron, label={[label distance=2mm]\ifthenelse{\y=1}{\bf{ReLU}}{}}] (F2-\name) at (\layersepedges + 1*\layersep + \forwardsep,-\vertSepFactory * \y cm) {};
            
    \foreach \name / \y in {1,...,2}
        \path[yshift=\shiftFactory]
            node[hidden neuron, text=white, label={[label distance=2mm]\ifthenelse{\y=1}{\bf{WS}}{}}] (B3-\name) at (\layersepedges + 2*\layersep,-\vertSepFactory * \y cm) {$+0$};

    \foreach \name / \y in {1,...,2}
        \path[yshift=\shiftFactory]
            node[hidden neuron, label={[label distance=2mm]\ifthenelse{\y=1}{\bf{ReLU}}{}}] (F3-\name) at (\layersepedges + 2*\layersep + \forwardsep,-\vertSepFactory * \y cm) {};

    \node[output neuron, label={[label distance=2mm]90:Output}]
        (O-1) at (\layersepedges*1 + 2*\layersep + 2cm + \forwardsep,-\vertSepFactoryI * 1 cm) {};

    \foreach \source in {1}
        \foreach \dest in {1,...,2}
            \draw[nnedge] (I-\source.east) -- node[above]{$1$}  (B1-\dest.west);
            
    \foreach \source in {1,...,2}
        \draw[nnedge] (B1-\source) -- (F1-\source);
        
    \draw[nnedge] (F1-1) -- node[above]{$1$} (B2-1.west);
    \draw[nnedge] (F1-1) -- node[above, near end]{$-1$} (B2-2.west);
    \draw[nnedge] (F1-2) -- node[above]{$1$} (B2-1.west);
    \draw[nnedge] (F1-2) -- node[below]{$1$} (B2-2.west);

    \foreach \source in {1,...,2}
        \draw[nnedge] (B2-\source) -- (F2-\source);
      
    \draw[nnedge] (F2-1) -- node[above]{$1$} (B3-1.west);
    \draw[nnedge] (F2-1) -- node[above, near end]{$1$} (B3-2.west);
    \draw[nnedge] (F2-2) -- node[above]{$1$} (B3-1.west);
    \draw[nnedge] (F2-2) -- node[below]{$1$} (B3-2.west);

    \foreach \source in {1,...,2}
        \draw[nnedge] (B3-\source) -- (F3-\source);
    
    \foreach \source in {1,...,2}
        \foreach \dest in {1}
            \draw[nnedge] (F3-\source.east) -- node[above]{$1$} (O-\dest.west);

\end{tikzpicture}
}

\caption {The \textcolor{color1}{\textbf{orange}} \relu{} neuron, marked $y$,
is \emph{2-forward-redundant}.
Replacing $y$ with a constant zero affects the following WS and ReLU layers,
but it does not affect the last WS layer (and thus the network output).
For example, observe that if we input $1$ into the network, $y$
evaluates to $1$, and the network's output evaluates to $12$. This
output value is unchanged even 
if we replace $y$'s value with $0$. A careful examination of the
network reveals that this will always be the case, regardless of the
network's input value.}
\label{fig:forward-red-example}
\end{figure}
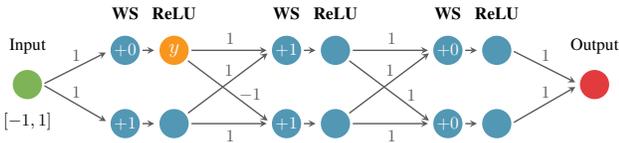

\medskip
\noindent
\textbf{Forward Redundancy.}  Phase-redundancy captures the case where
an activation function neuron is fixed to a single linear phase, for
all possible inputs. However, there actually exist \emph{unstable}
activation-function neurons, i.e. neurons not fixed to a particular
linear phase, which can still be soundly transformed into weighted-sum
neurons computing one of these linear phases. Intuitively, this
happens when neuron $y$'s assignment affects its $k$ succeeding
layers, for some $k>0$, but gets ``canceled out'' in layer $k+1$. A
small, illustrative example appears in
Fig.~\ref{fig:forward-red-example}.  When replacing $y$ with a
weighted-sum neuron only affects neurons that are at most $k$ layers
away from $y$, we say that $y$ is \emph{\kfr{}}. Much like
phase-redundant neurons, \kfr{} neurons can be removed from the
network without harming its accuracy.

More formally, let $v_i^j$ be an activation function neuron, and let
$N'$ be a network obtained from $N$ replacing $v$ with a weighted-sum neuron
$v_i^j = b_i^j + \sum c_kx_k$. Let $V_1$ denote an input vector, on which
both $N$ and $N'$ are evaluated; and let $V_2,\ldots, V_n$ and
$V_2',\ldots, V_n'$ denote the layer evaluations of $N$ and $N'$
(respectively) on
 $V_1$. If, for every $V_1$, it holds that
$V_{i+k}=V'_{i+k}$, then we say that neuron $v_i^j$ is \kfr{}
(note that this implies $V_{i+k'}=V'_{i+k'}$ for every $k'>k$).  We
note that a neuron that is phase-redundant is also \kfr{}, for any
$k\geq 0$.

\begin{figure}[htp]

\centering     
\subfigure[Replacing a \relu{} with the zero function]{\includegraphics[scale=0.25]{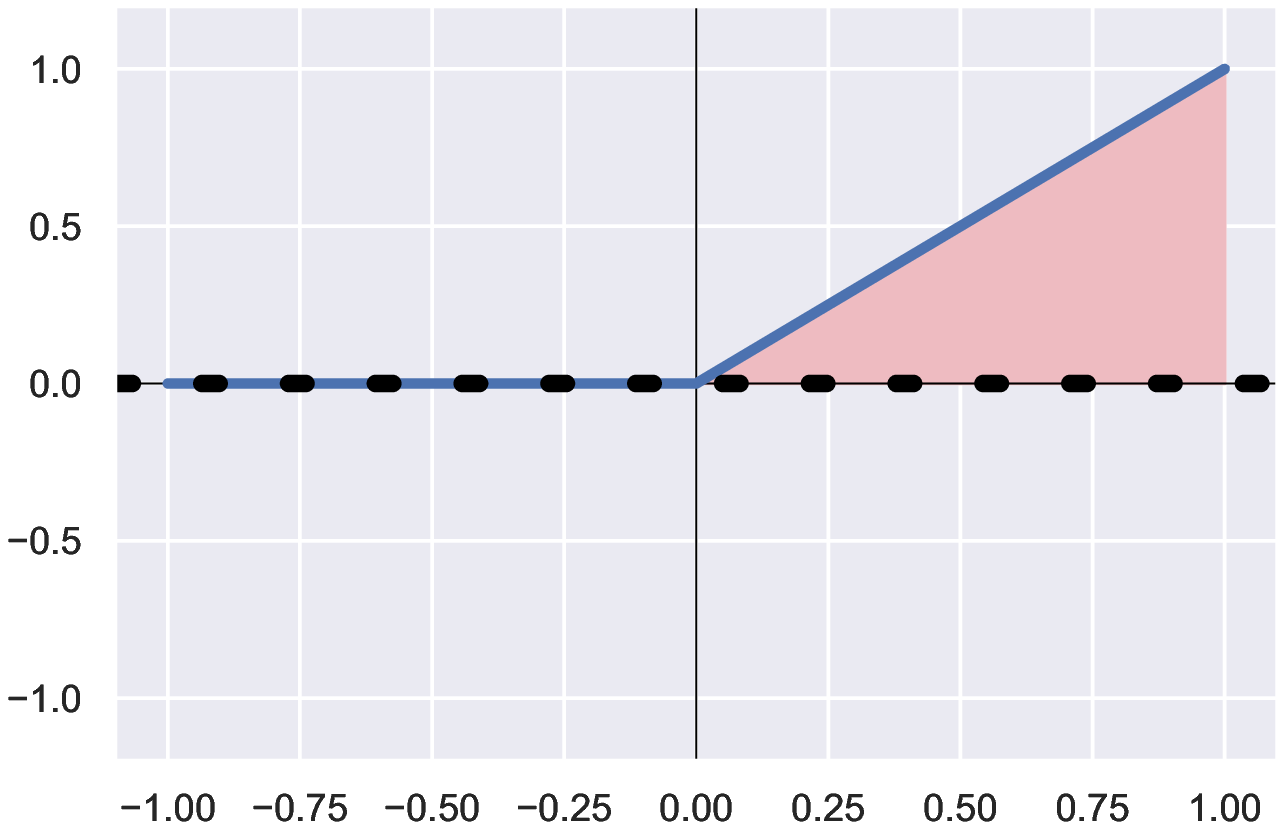}}
\subfigure[Replacing a \relu{} with identity function]{\includegraphics[scale=0.25]{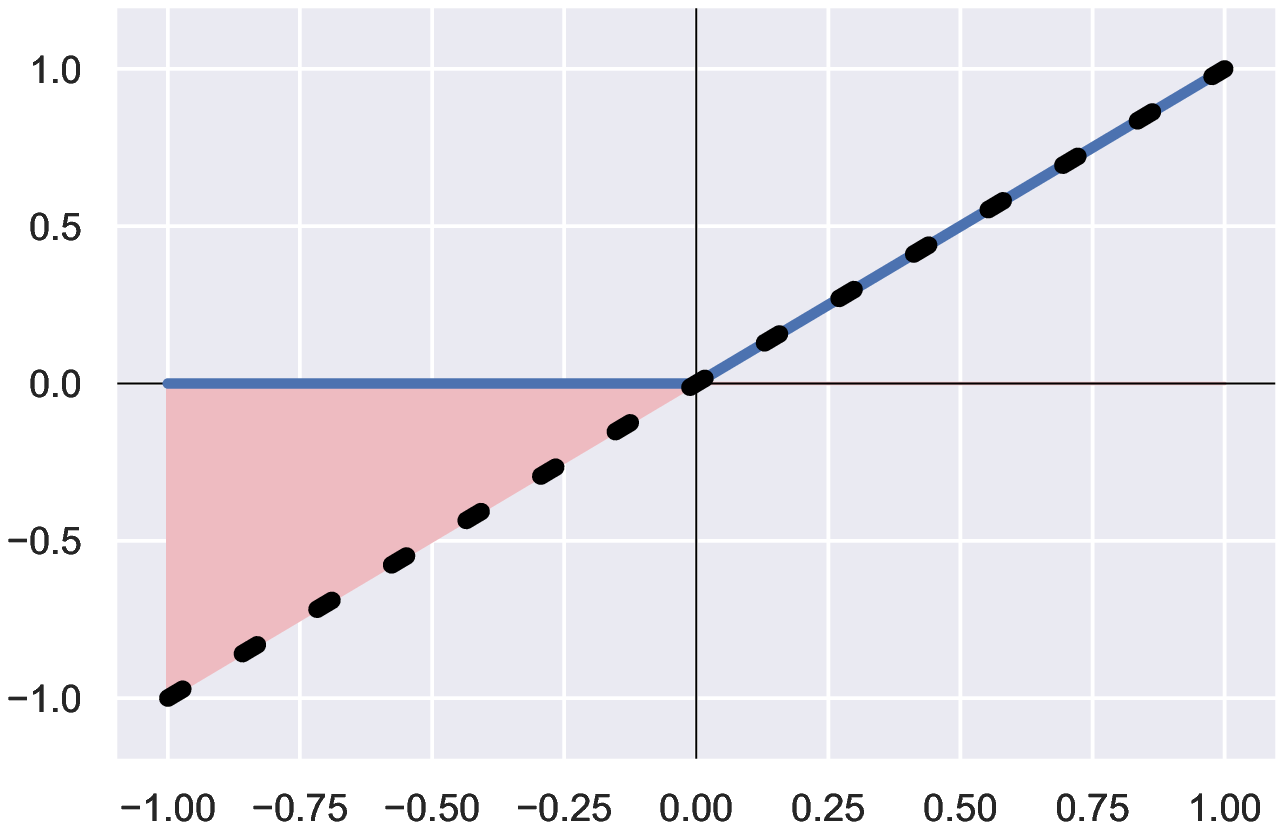}}
\subfigure[Replacing a \relu{} with an arbitrary linear function]{\includegraphics[scale=0.25]{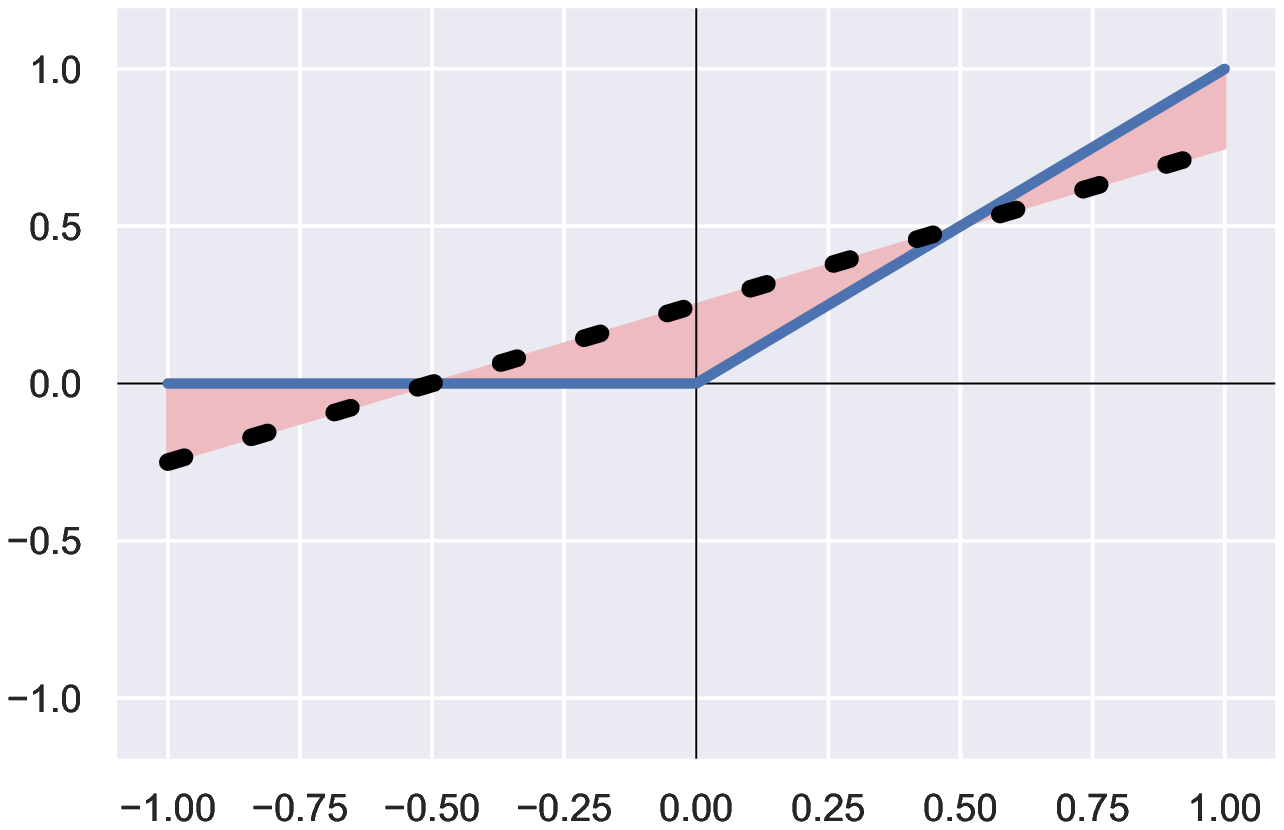}}

\caption {Replacing a \relu{} with linear functions.}
\label{fig:relaxed}
\end{figure}

\medskip
\noindent
\textbf{Relaxed Redundancy.}  So far, we discussed replacing a
piecewise-linear activation neuron with a weighted-sum neuron that
corresponds to one of the activation function's linear segments; e.g.,
in the case of $y=\relu{}(x)$, neuron $y$ would be changed into a
weighted-sum neuron computing either $y=0$ or $y=x$.  We observe that,
although these linear functions are natural candidates for replacing
the original constraint, in fact any linear function $y=\ell(x)$ could
be used. Specifically, given an activation function $y=f(x)$ and some
known lower and upper bounds $lb$ and $ub$ for $x$ (computed, e.g.,
using interval arithmetic~\cite{KaBaDiJuKo17} or abstract
interpretation~\cite{WaPeWhYaJa18,GeMiDrTsChVe18}), we propose to find
a linear function $\ell(x)$ that has \emph{minimal error} compared to
$f(x)$.  We define this error to be
\[
  \max_{lb\leq x\leq ub}|f(x)-\ell(x)|
  \]
See Fig.~\ref{fig:relaxed} for an
illustration of replacing a \relu{} constraint, whose phase is not fixed,
with three linear constraints. In each illustration, the blue line is
the \relu{}, the dashed line is the linear replacement, and the red
area is the introduced error. In case (c), the maximal introduced error (the
height of the red region) is the smallest among the three options.

Unlike in the phase-redundancy and \kfracy{} cases, setting
$y=\ell(x)$ will introduce some imprecision to the network's output.
The motivation is that by replacing $y=f(x)$ with $y=\ell(x)$ that has
minimal error, we would be introducing only a small imprecision, while enabling
the removal of $y$. Let $e_t$ be some user-defined error threshold;
when replacing $y=f(x)$ with $\ell(x)$ introduces an error $e$ such
that $e\leq e_t$, we say that neuron $y$ is \emph{relaxed-redundant}.

Let us focus on the $y=\relu{}(x)$ function as an example, and suppose
we know that $x\in [lb,ub]$. If $lb<0$ and $ub>0$, the neuron is not phase-redundant.
In this case, a linear function $y=\errorMinimizerFunction{}(x)$ with
minimal error can be easily computed, and is given by:
\[
  \errorMinimizerFunction{}(x) = \frac{ub}{ub - lb} \cdot x
  + \frac{-lb\cdot ub}{2(ub - lb)}.
\]
It is straightforward to check that the maximum error is obtained when
$x = 0$, and it is given by $\frac{-lb \cdot ub}{2(ub - lb)}$ (a proof
appears in Appendix~\ref{appendix:relax_analysis}).  Unsurprisingly,
when $lb$ or $ub$ are close to $0$, the error becomes very small ---
indicating that such \relu{}s, which are ``almost phase-redundant'',
could be removed at a small cost to precision.  It should be noted,
however, that minimizing the maximum error introduced by the removal
of a single neuron does not necessarily minimize the overall
imprecision introduced to the network's outputs.



\medskip
\noindent
\textbf{Result-Preserving Redundancy.}  In classification networks, it
may be acceptable to give up some precision, as long as the output
label for each input is unchanged; i.e., if the original network
classified input $x$ as label $l$ with $80\%$ confidence, it may be
acceptable to remove neurons in a way that reduces this confidence to
$60\%$, as long as $x$ is still classified as $l$.

\begin{figure}[htp]
\centering %
\scalebox{0.65}{%
\noindent\begin{tikzpicture}
\def\layersepedges{2.4cm}
\def\layersep{3.9cm}
\def\forwardsep{1.5cm}
\def\vertSepFactoryI{0.}
\def\vertSepFactory{1.3}
\def\shiftFactory{1.9cm}

    \node[input neuron, minimum size=0.9cm, label={[label distance=2mm]90:Input}, label={[label distance=1.5mm]270:$[-1,1]$}] (I-1) at (0,-\vertSepFactoryI * 1) {};

	\path[yshift=\shiftFactory]
            node[hidden neuron, minimum size=0.9cm, text=white, label={[label distance=2mm]\bf{WS}}] (B1-1) at (\layersepedges + 0*\layersep,-\vertSepFactory * 1 cm) {$+0$};
	\path[yshift=\shiftFactory]
            node[hidden neuron, minimum size=0.9cm, text=white] (B1-2) at (\layersepedges + 0*\layersep,-\vertSepFactory * 2 cm) {$-0.2$};

    \path[yshift=\shiftFactory]
            node[hidden neuron, minimum size=0.9cm, label={[label distance=2mm]\bf{ReLU}}] (F1-1) at (\layersepedges + 0*\layersep + \forwardsep,-\vertSepFactory * 1 cm) {};
\path[yshift=\shiftFactory]
            node[hidden neuron, minimum size=0.9cm, color=color2, text=white] (F1-2) at (\layersepedges + 0*\layersep + \forwardsep,-\vertSepFactory * 2 cm) {\large \bf $y$};
            
    \path[yshift=\shiftFactory]
            node[hidden neuron, minimum size=0.9cm, text=white, label={[label distance=2mm]{\bf{WS}}}] (B3-1) at (\layersepedges + 1*\layersep,-\vertSepFactory * 1 cm) {$+0$};
\path[yshift=\shiftFactory]
            node[hidden neuron, minimum size=0.9cm, text=white] (B3-2) at (\layersepedges + 1*\layersep,-\vertSepFactory * 2 cm) {$+0.1$};

    \foreach \name / \y in {1,...,2}
        \path[yshift=\shiftFactory]
            node[output neuron, minimum size=0.9cm, text=white, label={[label distance=1.5mm]\ifthenelse{\y=1}{Outputs}{}}] (F3-\name) at (\layersepedges + 7 + 1*\layersep + \forwardsep,-\vertSepFactory * \y cm) {$\#\y$};

    \foreach \source in {1}
        \foreach \dest in {1,...,2}
            \draw[nnedge] (I-\source.east) -- node[above]{$1$}  (B1-\dest.west);
            
    \foreach \source in {1,...,2}
        \draw[nnedge] (B1-\source) -- (F1-\source);
        
    \draw[nnedge] (F1-1) -- node[above]{$2$} (B3-1.west);
    \draw[nnedge] (F1-1) -- node[above, near end]{$1$} (B3-2.west);
    \draw[nnedge] (F1-2) -- node[above]{$1$} (B3-1.west);
    \draw[nnedge] (F1-2) -- node[below]{$-1$} (B3-2.west);

    \foreach \source in {1,...,2}
        \draw[nnedge] (B3-\source) -- (F3-\source);

\end{tikzpicture}
}
\caption {The \textcolor{color1}{\textbf{orange}} \relu{}, marked $y$, is
  result-preserving redundant and can be replaced with a constant zero.
  Observe that any input in range $(0.1, 1]$
  is classified as label $\#1$, while any input in range $[-1, 0.1)$ is
  classified as label $\#2$. The \relu{} in
  \textcolor{color1}{\textbf{orange}} is active only for inputs in
  $(0.2, 1]$, and it only increases the confidence in label $\#1$.
  For example, the network output for input $0.5$ is $[1.3, 0.3]^T$,
  and after replacing $y$ with $0$ the output becomes $[1.0, 0.6]^T$. Label $\#1$ still
  wins, but with a lower confidence.
  Thus, $y$ is result-preserving redundant --- replacing it with a
  constant zero does not change the winning class, for the entire
  input domain.}
\label{fig:respres-example}
\end{figure}
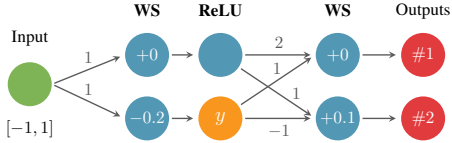

More formally, let $y=f(x)$ be an activation neuron in a network $N$,
and let $N'$ denote the same network with $y$ replaced by a weighted
sum neuron, $y=\ell(x)$. If, for every input vector $V_1$, it holds
that $\argmax(V_n)=\argmax(V'_n)$, i.e. if both networks classify each
input vector in the same way (regardless of the actual output neuron
values computed), then we say that neuron $y$ is
\emph{result-preserving redundant}. See Fig.~\ref{fig:respres-example}
for an example.

Note that result-preserving redundancy is, in a way, more permissive
than the previous categories: we do not directly try to bound the
imprecision introduced, but rather only try to maintain the same
output \emph{label} for every input.  Clearly, any neuron that is
phase-redundant or \kfr{} is also result-preserving; and it is
reasonable to assume that relaxed-redundant neurons with a small error
would also be result-preserving redundant. The motivation for
considering this kind of redundancy is that, due to its more
permissive nature, it can identify additional redundant neurons.

Our definition of result-preserving
redundancy can also be slightly relaxed, to exclude inputs whose
classification was \emph{borderline}; i.e., inputs whose
highest-scored label and the second-highest label received very
similar scores. Intuitively, with this alteration, a neuron is
considered result-preserving redundant if it does not change the
classification of any inputs which were previously classified with a
high degree of confidence, but may flip the classification of inputs
about which the DNN was not sure to begin with. The motivation for
this change is to allow the removal of additional neurons.

\section{Neuron Removal Strategies}
\label{sec:simultaneousRemoval}
In Section~\ref{sec:formalization} we laid the theoretical foundations
of our DNN simplification approach, by defining four kinds of
redundant neurons that could be removed to reduce network size. There
exist many strategies for applying these definitions in practice, in
order to reduce network sizes.  Intuitively, a good strategy is one
that identifies large sets of neurons that can be removed
simultaneously, in a way that is computationally efficient.  In this
Section, we propose one such strategy, which we have empirically
observed to perform well.

\medskip
\noindent
\textbf{Step 1: Bound Estimation using MILP.}
Let $v$ be an activation function neuron which we are considering for
removal. In this context, it is useful to deduce lower and upper
bounds for $v$ that are as tight as possible. Such bounds could lead,
for example, to the classification of $v$ as phase-redundant, or
enable us to compute $\errorMinimizerFunction(v)$ and declare $v$ to
be relaxed-redundant.

Mixed-Integer Linear Programming (MILP)~\cite{Dantzig1963} is a
well-studied method for solving a system of linear constraints with
real and integer variables. In the context of DNN verification, MILP
can be used to derive lower and upper bounds on the values that the
various neurons in the DNN can obtain~\cite{Ehlers2017,TjXiTe17}. This
is done by encoding a linear over-approximation of the neural network
into the MILP solver, and then using the solver's objective function
to maximize/minimize each of the individual neurons. For example,
after encoding a network $N$, we could set the solver's objective
function to $1\cdot v$, where $v$ is some neuron in $N$; and the
optimal solution discovered would then constitute $v$'s upper bound.

As a first step in the simplification process, we propose to run such
MILP queries for every neuron that is candidate for removal. The
number of resulting queries can be large --- two queries per neuron,
one for each bound --- but the gains are significant, as the
discovered bounds can often be quite tight~\cite{TjXiTe17}. At the end
of this step, we immediately remove all phase-redundant neurons.

In practice, it is useful to run the MILP solver with a short timeout
(e.g., 10 second)
for each neuron. In case a timeout occurs, modern
solvers are able to provide a sound approximation of the optimal
solution~\cite{Gurobi}.  In our experiments, we observed that this
initial step already detects a large number of phase-redundant
neurons.
%
%

\medskip
\noindent
\textbf{Step 2: Simulations.}  After the MILP phase is concluded, we
are left with multiple activation-function neurons whose phases are
not yet fixed. It is possible that some of these neurons are also
phase-redundant, but that the bounds discovered in the MILP pass were
too loose to indicate this. It is also possible that they are \kfr{}
or result-preserving redundant.  At this point we wish to quickly
\emph{rule out} as many of these candidates as possible, before
applying computationally expensive steps to dispatch the remaining
candidates.

To do this, we follow in the footsteps of Gokulanathan et
al.~\cite{GoFeMaBaKa20}, and apply \emph{simulations}; i.e., we
evaluate the network on a large number of random inputs, and for each
input record the values assigned to the network's neurons. Simulations
can easily show that a neuron is not phase-redundant, by demonstrating
two different inputs for which the neuron is in two different linear
phases. Similarly, they can show that a neuron is not \kfr{} or
result-preserving redundant.

\medskip
\noindent
\textbf{Step 3: Formal Verification.}  After the MILP and simulation
phases, we are left with activation-function neurons that are
candidates for removal, if we can prove them redundant. We now apply
formal verification to classify these remaining neurons. Specifically,
for each candidate neuron $v$, we:
\begin{inparaenum}[(i)]
\item apply verification to check whether $v$ is fixed to one if its
  linear phases, and is hence phase-redundant; and if not,
\item if $N$ is a classification network,
  apply verification to check whether $v$ is result-preserving
  redundant; else, if $N$ is a regression network, 
  apply verification to check whether $v$ is \kfr{}, for a value of $k$ that
  corresponds to the output layer.
\end{inparaenum}
Each of these conditions can be posed as a DNN verification query, as
described next. As soon as a neuron is marked redundant, it is
removed, and the process continues.

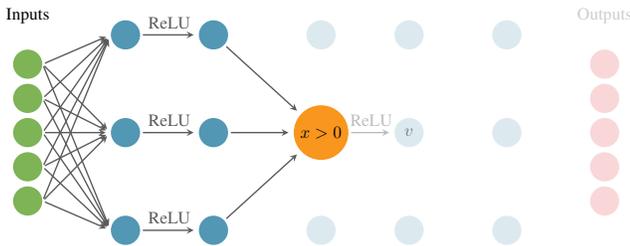
\begin{figure}[htp]
\centering%
\scalebox{0.65}{%
\begin{tikzpicture}
\def\layersepedges{2cm}
\def\layersep{4cm}
\def\forwardsep{1.8cm}
\def\vertSepFactoryI{0.7}
\def\vertSepFactory{2}
\def\shiftFactory{1.9cm}

    \foreach \name / \y in {1,...,5}
        \node[input neuron, label={[label distance=4mm]90:\ifthenelse{\y=1}{Inputs}{}}] (I-\name) at (0,-\vertSepFactoryI * \y) {};

    \foreach \name / \y in {1,...,3}
        \path[yshift=\shiftFactory]
            node[hidden neuron] (B1-\name) at (\layersepedges + 0*\layersep,-\vertSepFactory * \y cm) {};

    \foreach \name / \y in {1,...,3}
        \path[yshift=\shiftFactory]
            node[hidden neuron] (F1-\name) at (\layersepedges + 0*\layersep + \forwardsep,-\vertSepFactory * \y cm) {};

    \path[yshift=\shiftFactory]
            node[hidden neuron, color=color2, text=black] (B2-2) at (\layersepedges + 1*\layersep,-\vertSepFactory * 2 cm) {$\text{ }x > 0\text{ }$};

    \foreach \name / \y in {1,3}
        \path[opacity=0.2, yshift=\shiftFactory]
            node[hidden neuron] (B2-\name) at (\layersepedges + 1*\layersep,-\vertSepFactory * \y cm) {};

\foreach \name / \y in {1,...,3}
        \path[opacity=0.2, yshift=\shiftFactory]
            node[hidden neuron] (F2-\name) at (\layersepedges + 1*\layersep + \forwardsep,-\vertSepFactory * \y cm) {};
            
    \path[opacity=0.5, yshift=\shiftFactory]
            node (F2-2-title) at (\layersepedges + 1*\layersep + \forwardsep,-\vertSepFactory * 2 cm) {$v$};

    \foreach \name / \y in {1,...,3}
        \path[opacity=0.2, yshift=\shiftFactory]
            node[hidden neuron] (F3-\name) at (\layersepedges + 1*\layersep + 2cm + \forwardsep,-\vertSepFactory * \y cm) {};

    \foreach \name / \y in {1,...,5}
        \node[output neuron, opacity=0.2, label={[label distance=4mm, opacity=0.2]90:\ifthenelse{\y=1}{Outputs}{}}]
        (O-\name) at (\layersepedges*2 + 1*\layersep + 2cm + \forwardsep,-\vertSepFactoryI * \y cm) {};

    \foreach \source in {1,...,5}
        \foreach \dest in {1,...,3}
            \path (I-\source.east) edge[nnedge] (B1-\dest.west);
            
    \foreach \source in {1,...,3}
    	\path (B1-\source) edge[nnedge] node[above]{ReLU} (F1-\source);
    	
    \foreach \source in {1,...,3}
    	\path (F1-\source) edge[nnedge] (B2-2);
    	
    \path[opacity=0.4] (B2-2) edge[nnedge] node[above]{ReLU} (F2-2);

\end{tikzpicture}
}
\caption {A query for determining whether \relu{} node $v=\relu{}(x)$ is
  \emph{phase-redundant}: we check whether it is possible that
  $x>0$, and if not, we conclude that $v$ is inactive-redundant.
  To facilitate the verification process, the neurons in subsequent layers, as
  well as all other neurons in layer 2 (grayed out), are not encoded.}
\label{fig:inactive_query}
\end{figure}

In order to determine whether $v=f(x)$ is phase-redundant, we must
check whether $x$ is restricted to a certain linear segment. Let
$[s_1,s_2], [s_2,s_3], \ldots [s_k,s_{k+1}]$ be the set of possible
segments. For each such segment $[s_i,s_{i+1}]$, we can encode the DNN into the
verifier, and pose the query: $\exists V_1. (x<s_i) \vee (x>s_{i+1})$. If
the answer is \unsat{}, we know that $x$ is indeed fixed into segment 
$[s_i,s_{i+1}]$. An illustration appears in Fig.~\ref{fig:inactive_query}.

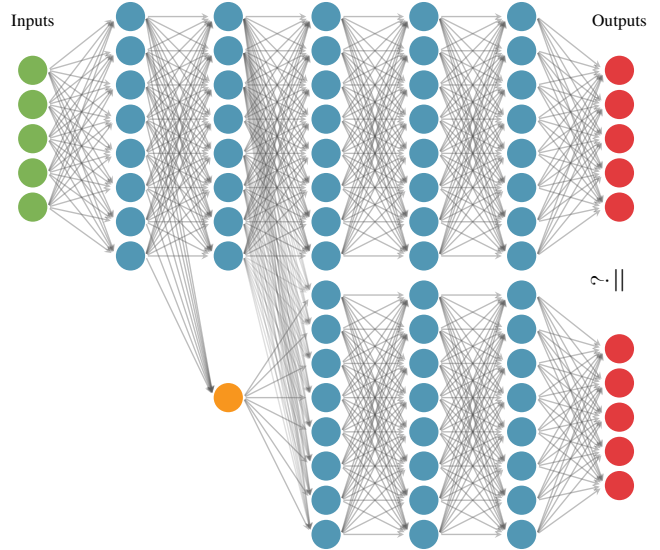
\begin{figure}[htp]
\centering%
\scalebox{0.65}{%
\begin{tikzpicture}
\def\layersepedges{2cm}
\def\layersep{4cm}
\def\forwardsep{2cm}
\def\vertSepFactoryI{0.7}
\def\vertSepFactory{0.7}
\def\shiftFactory{1.1cm}
\def\pathopacity{0.4}


    \foreach \name / \y in {4}
        \path[yshift=\shiftFactory]
            node[hidden neuron, color=color2, label={[label distance=0.6cm, color=OrangeRed, xshift=-0.35cm, text depth=-1ex, rotate=90]right:\ifthenelse{\y=1}{\huge \bf $\neq$}{}}] (F1-\name) at (\layersepedges + 0*\layersep + \forwardsep,-\vertSepFactory * \y cm) {};

    \foreach \name / \y in {1,...,8}
        \path[yshift=\shiftFactory]
            node[hidden neuron, label={[label distance=0.6cm, color=OrangeRed, xshift=-0.35cm, text depth=-1ex, rotate=90]right:\ifthenelse{\y=1}{}{}}] (B2-\name) at (\layersepedges + 1*\layersep,-\vertSepFactory * \y cm) {};

    \foreach \name / \y in {1,...,8}
        \path[yshift=\shiftFactory]
            node[hidden neuron, label={[label distance=0.6cm, color=OrangeRed, xshift=-0.35cm, text depth=-1ex, rotate=90]right:\ifthenelse{\y=1}{}{}}] (F2-\name) at (\layersepedges + 1*\layersep + \forwardsep,-\vertSepFactory * \y cm) {};

    \foreach \name / \y in {1,...,8}
        \path[yshift=\shiftFactory]
            node[hidden neuron, label={[label distance=0.6cm, color=LimeGreen, xshift=-0.25cm, text depth=-1ex, rotate=90]right:\ifthenelse{\y=1}{}{}}] (F3-\name) at (\layersepedges + 1*\layersep + 2cm + \forwardsep,-\vertSepFactory * \y cm) {};


    \foreach \name / \y in {1,...,5}
        \node[output neuron, label={[xshift=0.3 cm, yshift=1.1 cm, rotate=90]\ifthenelse{\y=1}{\huge \bf $\stackrel{?}{=}$}{}}]
        (O-\name) at (\layersepedges*2 + 1*\layersep + 2cm + \forwardsep,-\vertSepFactoryI * \y cm) {};

            
    \foreach \source in {1,...,8}
        \foreach \dest in {1,...,5}
            \path[opacity=\pathopacity] (F3-\source.east) edge[nnedge] (O-\dest.west);

    \foreach \source in {4}
        \foreach \dest in {1,...,8}
            \path[opacity=\pathopacity] (F1-\source.east) edge[nnedge] (B2-\dest.west);

    \foreach \source in {1,...,8}
        \foreach \dest in {1,...,8}
            \path[opacity=\pathopacity] (B2-\source.east) edge[nnedge] (F2-\dest.west);
            
    \foreach \source in {1,...,8}
        \foreach \dest in {1,...,8}
            \path[opacity=\pathopacity] (F2-\source.east) edge[nnedge] (F3-\dest.west);

	\begin{scope}[yshift={5.7cm}]

    \foreach \name / \y in {1,...,5}
        \node[input neuron, label={[label distance=4mm]90:\ifthenelse{\y=1}{Inputs}{}}] (I-\name) at (0,-\vertSepFactoryI * \y) {};

    \foreach \name / \y in {1,...,8}
        \path[yshift=\shiftFactory]
            node[hidden neuron] (SB1-\name) at (\layersepedges + 0*\layersep,-\vertSepFactory * \y cm) {};

    \foreach \name / \y in {1,...,8}
        \path[yshift=\shiftFactory]
            node[hidden neuron] (SF1-\name) at (\layersepedges + 0*\layersep + \forwardsep,-\vertSepFactory * \y cm) {};

    \foreach \name / \y in {1,...,8}
        \path[yshift=\shiftFactory]
            node[hidden neuron] (SB2-\name) at (\layersepedges + 1*\layersep,-\vertSepFactory * \y cm) {};

    \foreach \name / \y in {1,...,8}
        \path[yshift=\shiftFactory]
            node[hidden neuron] (SF2-\name) at (\layersepedges + 1*\layersep + \forwardsep,-\vertSepFactory * \y cm) {};

    \foreach \name / \y in {1,...,8}
        \path[yshift=\shiftFactory]
            node[hidden neuron] (SF3-\name) at (\layersepedges + 1*\layersep + 2cm + \forwardsep,-\vertSepFactory * \y cm) {};

    \foreach \name / \y in {1,...,5}
        \node[output neuron, label={[label distance=4mm]90:\ifthenelse{\y=1}{Outputs}{}}]
        (O-\name) at (\layersepedges*2 + 1*\layersep + 2cm + \forwardsep,-\vertSepFactoryI * \y cm) {};
    \foreach \source in {1,...,5}
        \foreach \dest in {1,...,8}
            \path[opacity=\pathopacity] (I-\source.east) edge[nnedge]  (SB1-\dest.west);
            
    \foreach \source in {1,...,8}
        \foreach \dest in {1,...,5}
            \path[opacity=\pathopacity] (SF3-\source.east) edge[nnedge] (O-\dest.west);

    \foreach \source in {1,...,8}
        \foreach \dest in {1,...,8}
            \path[opacity=\pathopacity] (SF1-\source.east) edge[nnedge] (SB2-\dest.west);

    \foreach \source in {1,...,8}
        \foreach \dest in {1,...,8}
            \path[opacity=\pathopacity] (SB1-\source.east) edge[nnedge] (SF1-\dest.west);
            
    \foreach \source in {1,...,8}
        \foreach \dest in {1,...,8}
            \path[opacity=\pathopacity] (SB2-\source.east) edge[nnedge] (SF2-\dest.west);
            
    \foreach \source in {1,...,8}
        \foreach \dest in {1,...,8}
            \path[opacity=\pathopacity] (SF2-\source.east) edge[nnedge] (SF3-\dest.west);

\end{scope}

    \foreach \source in {1,2,3,5,6,7,8}
        \foreach \dest in {1,...,8}
            \path[opacity=0.2] (SF1-\source.east) edge[nnedge] (B2-\dest.west);
            
    \foreach \source in {1,...,8}
        \foreach \dest in {4}
            \path[opacity=\pathopacity] (SB1-\source.east) edge[nnedge] (F1-\dest.west);

\end{tikzpicture}
}
\caption {{\bf 4-Forward-Redundancy} query illustration. The neuron in
\textcolor{color1}{\textbf{orange}} is the neuron being checked for forward-redundancy.
In this case the layer being checked is at distance 4, which happens to be the output
layer.
\label{fig:forward-query}}
\end{figure}

Determining whether $v=f(x)$  is \kfr{} is done by
creating a query where the part of the network starting from the
neuron in question is duplicated. One copy of the network is the
unmodified one, and in the other copy $v=f(x)$ is replaced with a
weighted-sum neuron, $v'=\ell(x)$.   We query the verifier whether it
is possible that a neuron $k$ layers away from $v$ is assigned 
different values in the original and modified copies. If the answer is
\unsat{}, the neuron is \kfr{}. See Fig.~\ref{fig:forward-query}
for an illustration.

Determining whether $v=f(x)$ is result-preserving redundant
is done by creating a query similar to the \kfr{} case, only this time
we ask the verifier whether there exists
an input that the two networks classify differently. If the answer is
\unsat{}, we know that the neuron is indeed result-preserving
redundant.

\medskip
\noindent
\textbf{Step 4: Relaxed Redundancy and Accumulative Error.}  The
aforementioned steps were aimed at identifying and removing redundant
neurons, without introducing any imprecision into the simplified
network. Last but not least, we discuss the removal of
relaxed-redundant neurons. Recall that relaxed-redundant neurons are
determined by a user-specified error threshold $e_t$.  Identifying
these neurons is thus a local operation, that does not require
verification; for every neuron we can compute the maximum error
introduced by replacing it with $\errorMinimizerFunction$, and see
whether it exceeds the threshold.

While each relaxed-redundant neuron can be identified locally,
removing multiple neurons simultaneously runs the risk of compounding
the overall error, beyond the permitted threshold.  To circumvent this
issue and allow the efficient removal of multiple relaxed-redundant
neurons, we introduce the following lemma:

\begin{lemma}
  Let $N$ be a neural network, and let $N'$ be a simplified  network,
  obtained from $N$ by removing relaxed-redundant neurons $u_1,\ldots,u_n$. Consider
  another neuron $v$ in $N'$ that is relaxed-redundant, and let $e_{in}$
  denote the error to $v$'s \emph{input}, previously introduced by the removal of
  $u_1,\ldots,u_n$. Let $e_v$ denote the error introduced by the
  removal of $v$. Then, if we remove $v$, the overall error introduced
  to its \emph{output} is upper bounded by:
\[
	e_{in} + e_v
\]
\end{lemma}

This lemma tells us that the iterative removal of relaxed-redundant
neurons does not compound the introduced error; instead, the error
introduced by the removal of each neuron is only added to the error
already introduced by the removal of other neurons.
This enables us, through a straightforward computation, to upper bound
the overall imprecision (on the output layer) that the removal of a
set of relaxed-redundant neurons might cause. Consequently, our
proposed strategy is to begin removing relaxed-redundant neurons with
small error rates, each time recomputing the overall network
inaccuracy, until hitting the prescribed overall error threshold.  A
full, formal description of these claims appears in
Appendix~\ref{appendix:sim_error_bounds}.

\section{Introducing Redundancies via Input Slicing}
\label{sec:input-slicing}
So far, our simplification efforts have hinged on the existence of
redundant neurons. Next, we introduce a technique that can cause
neurons to become redundant, even if they are initially not so.

The core idea is to:
\begin{inparaenum}[(i)]
  \item \emph{slice the input domain} $\mathcal{D}$ of the DNN $N$ into smaller
    sub-domains $\mathcal{D}_1,\ldots, \mathcal{D}_n$;
  \item duplicate the original network $n$ times, resulting in
    networks $N_1,\ldots,N_n$, such that network $N_i$ is associated
    with domain $\mathcal{D}_i$; and
  \item apply the simplification process
    described in
    Section~\ref{sec:simultaneousRemoval} for each $N_i$, separately. 
  \end{inparaenum}
  Intuitively, splitting the input domain into sub-domains can serve
  to separate ``simpler'' inputs regions, in which many neurons are
  phase-redundant, from more ``complex'' input domains where neurons
  fluctuate between phases. Various heuristics can be used for splitting
  the input domain, depending on the network in question. 
  A simple splitting method, which we used in our evaluation, is to
  split the range of each input coordinate into $n$ even sub-ranges.
  
  %

  After the slicing and simplification is done, we are left with a
  family of DNNs $N_1,\ldots,N_n$, which are together equivalent to
  the original $N$.  Evaluation is then performed in two steps: given
  an input vector $V_1$, we first identify the domain $\mathcal{D}_i$
  to which $V_1$ belongs; and then compute $N_i(V_1)$ and return the
  result. As our evaluation shows, the resulting $N_i$ networks can be
  quite small, resulting in a significant improvement to the expected
  number of operations required for evaluating the network. This
  improvement might come at the expense of increased space
  requirements for storing the resulting family of networks, making
  this approach suitable for cases where space is abundant but fast
  inference is crucial.  We note that, as a side effect, the resulting
  networks may be easier to
  verify~\cite{WaPeWhYaJa18,WuOzZeIrJuGoFoKaPaBa20}.

\medskip
\noindent
\textbf{Discussion: Dependency on Input Dimensions.}
Our proposed slicing method relies on splitting the input domain, by
restricting input neurons to various values. This approach works quite
well on DNNs with relatively few input neurons (e.g., the ACAS Xu
family of networks~\cite{JuLoBrOwKo16}; see
Section~\ref{sec:evaluation} for details). For networks with a
larger number of input neurons (e.g., image recognition networks),
the number of input sub-domains might be prohibitively large. Indeed, a
similar phenomenon has been observed for verification techniques that
rely on input slicing~\cite{WaPeWhYaJa18,WuOzZeIrJuGoFoKaPaBa20}.

One approach for mitigating this difficulty is through performing
slicing not on the input layer, but on some smaller intermediate layer $L_k$
in the network. Then, the network would be evaluated by evaluating the
original network's layers $L_1\ldots L_{k-1}$, and then using the
values computed for layer $L_k$ in choosing from a set of networks for
continuing the evaluation. We speculate that for an intermediate layer
of a moderate size, this approach could lead to improved performance
over input slicing. We leave this for future work.



\medskip
\noindent
\textbf{Extreme Slicing: Complete Linearization.}  We observe that
input slicing can be used to completely linearize every sub-domain of
the input space; that is, if the resulting sub-domains are
sufficiently small, then in each network $N_i$ all activation
functions will become phase-redundant, effectively collapsing the DNN
into a linear transformation.  Additionally, even if the slicing does
not fix the phase of all activation function neurons, extreme slicing
tends to decrease the error introduced by removing relaxed-redundant
neurons; and thus, complete linearization could be achieved by
removing these neurons, even if they have not become phase-redundant.
This linearization approach can thus be regarded as providing us with
a simple, piecewise-linear approximation of the network as a whole ---
with an upper bound on the error in each sub-domain.  Our experimental
results in Section~\ref{sec:evaluation} demonstrate very low error
rates on most sub-domains.

Complete linearization incorporates a trade-off: in order to obtain
very small, nearly-linear networks, the input domain would have to be
sliced many times. Users can fine-tune the number of slices used, and
consequently the sizes of the resulting DNNs, to their specific needs.





\section{Evaluation}
\label{sec:evaluation}
We created a proof-of-concept implementation of our approach as a
Python framework, available online~\cite{ourCode} (together with all
benchmarks reported in this section). The framework provides all the
functionality discussed so far: after importing a network, it can run
MILP queries to compute neuron bounds; perform simulations; and
identify phase-redundant, \kfr{} and result-preserving redundant
neurons, by running verification queries. The framework uses
the Gurobi~\cite{Gurobi} MILP solver and the Marabou~\cite{Marabou2019} DNN
verification engine as backends, although other backends could also be
used.

For evaluation purposes, we conducted extensive experiments on the
ACAS Xu system: an airborne collision avoidance system, implemented as
a family of 45 neural networks~\cite{JuLoBrOwKo16}. Each of these
neural networks has 5 input neurons, 5 output neurons, and 6 hidden
layers with 50 neurons each and ReLU activation functions (310 neurons
in total). Keeping the network sizes small was a key consideration in
developing the ACAS Xu system~\cite{JuLoBrOwKo16}, making it a prime
candidate on which to apply simplification techniques.

We began by comparing our approach to that of
Gokulanathan et al.~\cite{GoFeMaBaKa20}, which is the current
state-of-the-art in verification-based simplification of DNNs. Their
technique can be regarded as a private-case of ours, in which only
specific phase-redundant neurons (specifically, inactive-redundant
ReLUs) are removed. We compared that approach to our framework,
configured to identify and remove both active-redundant and
inactive-redundant ReLUs, and also to remove relaxed-redundant neurons.
We ran both tools on all 45 ACAS Xu networks; the
results appear in Table~\ref{table:sumathi}.

\newcommand{\redbymilp}{10.09}
\newcommand{\redbyformal}{2.42}
\newcommand{\inactivered}{12.09}
\newcommand{\activered}{0.42}
\newcommand{\relaxedfour}{0.22}
\newcommand{\relaxedthree}{1.22}
\newcommand{\relaxedtwo}{2.18}
\newcommand{\totalred}{\inactivered+\activered+\relaxedtwo}

\begin{table}[htp]
  \centering
  \caption{Phase-Redundancy and Relaxed-Redundancy on ACAS Xu networks.}
  \scalebox{0.88}{
    \begin{tabular}{ |c||c|c|c|c|c| } 
 \hline
      & Inactive & Active & \multicolumn{3}{c|}{Relaxed-Redundant} \\
      \cline{4-6}
      & Redundant & Redundant & $\epsilon=10^{-4}$ & $\epsilon=10^{-3}$ & $\epsilon=10^{-2}$ \\
 \hline
 
 
 \thead{\% of all\\ neurons} &
 $\fpeval{round(\inactivered / 300. * 100, 1)}\%$ &
 $\fpeval{round((\inactivered+\activered) / 300. * 100, 1)}\%$ &
 $\fpeval{round((\inactivered+\activered+\relaxedfour) / 300. * 100, 1)}\%$ &
 $\fpeval{round((\inactivered+\activered+\relaxedthree) / 300. * 100, 1)}\%$ &
 $\fpeval{round((\inactivered+\activered+\relaxedtwo) / 300. * 100, 1)}\%$
 \\ 

 \hline

 \thead{\% of \\redundant\\ neurons} &
 baseline &
 $\fpeval{round((\activered) / (\inactivered) * 100, 1)}\%$ &
 $\fpeval{round((\activered+\relaxedfour) / (\inactivered) * 100, 1)}\%$ &
 $\fpeval{round((\activered+\relaxedthree) / (\inactivered) * 100, 1)}\%$ &
 $\fpeval{round((\activered+\relaxedtwo) / (\inactivered) * 100, 1)}\%$
 \\ 
 
 \hline
 	\thead{output error\\ bound} &
 	0 & 0 & 0.02 & 2.64 & 525.1
  \\
  \hline
 	
    \end{tabular}
  }
  \label{table:sumathi}
\end{table}

The table depicts the accumulated numbers of redundant neurons, when
read from left to right (which is the order in which the techniques
were applied). First, inactive-redundant neurons are removed (this is
the technique of~\cite{GoFeMaBaKa20}), accounting for $4$\% of all
neurons in the network.
Active-redundant neurons are next, removing another $0.2$\%
of all neurons, which is a $3.5$\% increase in the number of removed
neurons. Finally, relaxed-redundant neurons are removed, with three
possible alternative $\epsilon$ values. The most permissive one,
$\epsilon=10^{-2}$, leads to the removal of $4.9$\% of the neurons in
total, which is a $21.5$\% increase over the baseline --- but the
resulting network error bound in this case, $525.1$, is quite high.
$\epsilon=10^{-3}$ appears a better choice, with a total removal rate
of $4.6$\% and a significantly smaller error bound of $2.64$.  We note
that our evaluation indicates that the output error bounds currently
computed are far from tight; devising tighter bounding schemes is a
work in progress.





In our second experiment, we evaluated our complete simplification
pipeline. First, we applied input-slicing, dividing the input domain
into 32,768 equal sub-domains (3 rounds of bisecting the range of each
of the $5$ input neurons in 2). Next, for each sub-domain we:
\begin{inparaenum}[(i)]
\item ran MILP and removed any discovered phase-redundant neurons;
\item ran simulations, and then formal verification to discover and
  remove any remaining phase-redundant neurons; and
\item identified all result-preserving neurons, and greedily attempted to
  simultaneously remove large sets thereof, using verification.
\end{inparaenum}
We note that identifying the largest set possible of result-preserving neurons that
can be removed simultaneously is a difficult problem, and our current
heuristic was a simple, greedy approach. Devising more sophisticated
heuristics is left for future work.

\newcommand{\statsMilp}{68.5}
\newcommand{\statsNonRedBySim}{26.4}
\newcommand{\statsRedByFormal}{1.7}
\newcommand{\statsResPres}{12.3}
\newcommand{\statsUnknowns}{7.2}
\newcommand{\statsTotalRed}{\fpeval{\statsMilp+\statsRedByFormal+\statsResPres}}

\begin{figure}[htp]
\begin{center}
\includegraphics[scale=0.5]{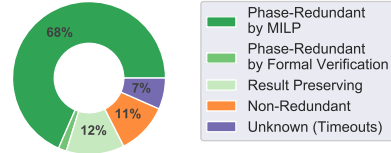}
\end{center}
\caption {Redundant neuron removal, averaged over 10 ACAS Xu input
  sub-domains.}
\label{fig:evaluation_acas}
\end{figure}

We ran the MILP step on all 32,768 sub-domains, which resulted in the
discovery of 67.3\%
phase-redundant neurons on average in each sub-domain.
We continued to run the pipeline on a
sample of 50 sub-domains selected at random.
Most notably, we observed an average removal of \emph{
  \statsTotalRed\% redundant neurons} (out of all neurons in the
network), with \statsUnknowns\% additional neurons still candidates
for removal, but for which the underlying verification engine
timed-out. Of the \statsTotalRed\% removed neurons, \fpeval{\statsMilp + \statsRedByFormal}\% were
phase-redundant, which is a very significant increase from the 4.2\% neurons
removed when the pipeline was run over the entire input domain. This
demonstrates the high effectiveness of input slicing. In addition,
about 21\% of phase-redundant neurons were active-redundant, which
signifies the importance of the generalization from ``dead neurons''~\cite{GoFeMaBaKa20} to
phase-redundancy. The remaining
\statsResPres\% neurons removed were result-preserving redundant.
Fig.~\ref{fig:evaluation_acas} shows the breakdown.

Slicing is highly beneficial for neuron removal, but results in a
large number of sub-domains that need to be checked. Within our
pipeline, verification steps are the most expensive, whereas MILP
queries and simulations are relatively cheap. We observe, however,
that MILP queries already account for most of the removed neurons.
Specifically, \statsMilp\% of all phase-redundant neurons removed were
discovered through MILP (about \fpeval{round((\statsMilp /
  \statsTotalRed)*100, 1)}\% of all redundant neurons), with a 10 second
timeout for each individual MILP query. 

The next step, namely simulations, is also computationally cheap and
highly effective.  For each sub-domain, we ran 100,000 simulations;
and out of the of \fpeval{100 - \statsMilp}\% neurons which were still
candidates for removal after the MILP phase, an average of
\statsNonRedBySim\% of the neurons were ruled not phase-redundant
through simulations. This left only a small number of candidates to be
dispatched through verification (\fpeval{100 - \statsMilp -
  \statsNonRedBySim}\% of the neurons), which in turn discovered the
remaining \statsRedByFormal\% redundant neurons, on average. In our
experiment, each Marabou verification query was run with a 4-hour timeout.

As discussed above, we used a fairly na\"ive strategy for discovering
result-preserving redundant neurons. Specifically, we ran formal
verification on each candidate neuron to check whether it was 
individually result-preserving redundant; this resulted in a set of 
candidates for removal.  Then, we ran result-preserving simulations, iteratively
removing additional candidate neurons from the network, as
long as the simulations could
not find a counter-example to the redundancy of the currently removed
set.
Finally, we ran a single verification query to verify that removing
our selected neurons was indeed a result-preserving operation.
On 75\% of the
sub-domains checked, this strategy worked. In sub-domains where we were
successful, we found an additional \fpeval{round(\statsResPres*2,
  1)}\% forward-redundant and result-preserving redundant neurons;
whereas in sub-domains where we were not successful, we had
a similar amount of candidates for removal on average.

\begin{figure}[htp]
\begin{center}
\includegraphics[scale=0.5]{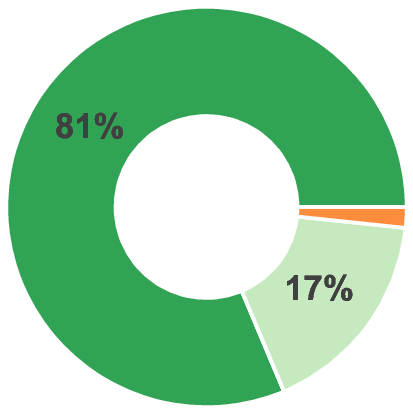}
\includegraphics[scale=0.5]{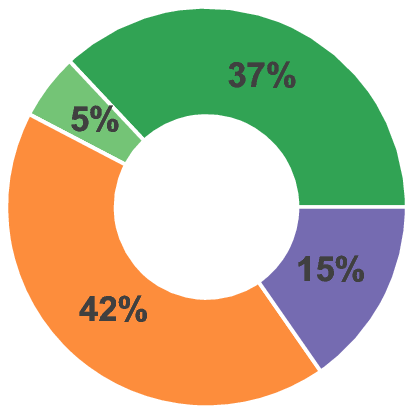}
\includegraphics[scale=0.6]{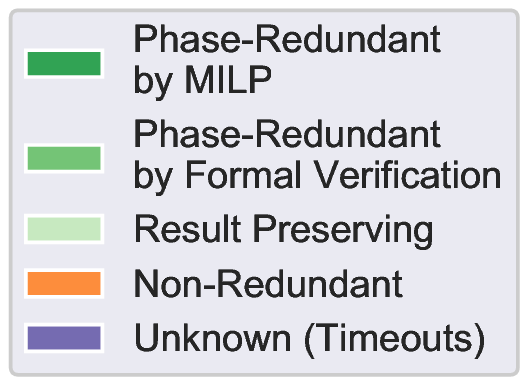}
\end{center}
\caption {An ``almost'' linear sub-domain (left) vs. a complex sub-domain (right).}
\label{fig:subspace_comp}
\end{figure}

In the final step of our experiment, we tested our hypothesis that
slicing can lead to the complete linearization of some of the
sub-domains. Indeed, for some of the sub-domains explored, the
simplification pipeline was able to remove \emph{all} neurons,
resulting in a DNN that is effectively a linear transformation. We
noticed, however, a high variability --- for example, in another
sub-domain we were only able to remove 58\% of the neurons.  See
Fig.~\ref{fig:subspace_comp} for additional details. We conclude that
there is an inherent difference between the sub-domains: apparently,
some of them compute simpler transformations than others.

\section{Related Work}
\label{sec:relatedWork}

The pruning of DNNs in order to reduce their sizes has received
significant attention from the machine learning community in recent
years. The most common approaches are based on heuristically
identifying neurons and edges that seem to contribute little to the
network's output, removing these neurons and edges, and performing
additional training of the
network~\cite{HaMaDa15,IaHaMoAsDaKe16}. Other approaches apply
quantization: by using fewer bits to store the network's weights or
activation functions, the DNN's footprint is
decreased~\cite{HuCoSoElBe17,RaOrReFa16,HuCoSoElBe16}. A common trait
of these approaches is that, while they achieve a significant reduction
in memory, they provide no guarantees about the resemblance of the
smaller network to the original.

The most closely related work to our own is that of Gokulanathan et
al.~\cite{GoFeMaBaKa20}. There, the authors use formal verification to
remove dead neurons from a network, ensuring that the resulting
network is equivalent to the original. Additionally, simulations are
used to reduce the number of verification queries that need to be
dispatched. Our work uses similar principles, but significantly
extends them: we consider additional kinds of redundancy
(phase-redundancy, \kfr{}, and result-preserving redundancy) that
produce equivalent networks, and also relaxed-redundancy which
removes additional neurons by introducing a bounded amount of
imprecision.

Our work uses the Marabou DNN verification engine as a
backend~\cite{Marabou2019,StWuZeJuKaBaKo20,KaBaDiJuKo17Fvav,CaKaBaDi17,
GoKaPaBa18,ElKaKaSc21,AmScKa21,KaBaKaSc19};
 but any of the many approaches and tools
that have been proposed in recent years could be used as well. These
approaches leverage SMT solvers (e.g.,~\cite{HuKwWaWu17}), based on LP
and MILP solvers (e.g.,~\cite{LoMa17,Eh17,TjXiTe17, BuTuToKoMu18}),
the propagation of symbolic intervals and abstract interpretation
(e.g.,~\cite{WaPeWhYaJa18, GeMiDrTsChVe18, WeZhChSoHsBoDhDa18,
  TrBkJo20}), abstraction-refinement techniques (e.g.,~\cite{ElGoKa20,
  AsHaKrMu20}), and many others. Recent work has extended beyond
answering yes/no questions about DNNs, targeting tasks such as
automated DNN repair~\cite{KoLoJaBl20,GoAdKeKa20} and quantitative
verification~\cite{BaShShiMeSa19}.  Verification approaches have also
been proposed for recurrent networks~\cite{ZhShGuGuLeNa20,JaBaKa20},
which could potentially also be simplified.  As DNN verification
technology improves, the scalability of our approach will also
increase.

\section{Conclusion and Future Work}
\label{sec:conclusion}
Neural networks often suffer from a high degree of redundancy, which
affects evaluation time, memory footprint and verification costs. In
this paper we presented a novel technique to identify and remove such
redundancy. Our framework is customizable, allowing users to safely
trade network precision for size reduction, while maintaining the
introduced imprecision within a prescribed bound.

In the future, we plan to extend our work along multiple
axes. Specifically, we plan to research more intelligent techniques
for input domain slicing than coordinate-splitting; and also
compositional techniques that would allow us to split the network into
several sub-networks, identify redundancies in each of them, and then
re-combine the pruned network into a single network that is smaller
than the original. In addition, we plan to explore ways of combining
our pruning techniques with techniques from the related field of 
Boolean circuit simplification~\cite{ChMaCh96}.

\medskip
\noindent
\textbf{Acknowledgements.} We thank Ittai Rubinstein and Haoze Wu for
their contributions to this project.  The project was partially
supported by the Israel Science Foundation (grant number 683/18) and
the Binational Science Foundation (grant number 2017662).

{
\bibliographystyle{abbrv}
\bibliography{redundancy}

\begin{thebibliography}{10}

\bibitem{AmScKa21}
G.~Amir, M.~Schapira, and G.~Katz.
\newblock {Towards Scalable Verification of Deep Reinforcement Learning}.
\newblock In {\em Proc. 21st Int. Conf. on Formal Methods in Computer-Aided
  Design (FMCAD)}, 2021.

\bibitem{AmWuBaKa21}
G.~Amir, H.~Wu, C.~Barrett, and G.~Katz.
\newblock {An SMT-Based Approach for Verifying Binarized Neural Networks}.
\newblock In {\em Proc. 27th Int. Conf. on Tools and Algorithms for the
  Construction and Analysis of Systems (TACAS)}, pages 203--222, 2021.

\bibitem{AsHaKrMu20}
P.~Ashok, V.~Hashemi, J.~Kretinsky, and S.~M\"{u}hlberger.
\newblock {DeepAbstract: Neural Network Abstraction for Accelerating
  Verification}.
\newblock In {\em Proc. 18th Int. Symposium on Automated Technology for
  Verification and Analysis (ATVA)}, 2020.

\bibitem{BaShShiMeSa19}
T.~Baluta, S.~Shen, S.~Shinde, K.~Meel, and P.~Saxena.
\newblock {Quantitative Verification of Neural Networks and its Security
  Applications}.
\newblock In {\em Proc. ACM SIGSAC Conf. on Computer and Communications
  Security (CCS)}, pages 1249--1264, 2019.

\bibitem{BoDeDwFiFlGoJaMoMuZhZhZhZi16}
M.~Bojarski, D.~Del~Testa, D.~Dworakowski, B.~Firner, B.~Flepp, P.~Goyal,
  L.~Jackel, M.~Monfort, U.~Muller, J.~Zhang, X.~Zhang, J.~Zhao, and K.~Zieba.
\newblock {End to End Learning for Self-Driving Cars}, 2016.
\newblock Technical Report. \url{http://arxiv.org/abs/1604.07316}.

\bibitem{BuTuToKoMu18}
R.~Bunel, I.~Turkaslan, P.~Torr, P.~Kohli, and P.~Mudigonda.
\newblock {A Unified View of Piecewise Linear Neural Network Verification}.
\newblock In {\em Proc. 32nd Conf. on Neural Information Processing Systems
  (NeurIPS)}, pages 4795--4804, 2018.

\bibitem{CaKaBaDi17}
N.~Carlini, G.~Katz, C.~Barrett, and D.~Dill.
\newblock {Provably Minimally-Distorted Adversarial Examples}, 2017.
\newblock Technical Report. \url{https://arxiv.org/abs/1709.10207}.

\bibitem{ChMaCh96}
S.-C. Chang, M.~Marek-Sadowska, and K.-T. Cheng.
\newblock {Perturb and Simplify: Multilevel Boolean Network Optimizer}.
\newblock {\em IEEE Transactions on Computer-Aided Design of Integrated
  Circuits and Systems}, 15(12):1494--1504, 1996.

\bibitem{Dantzig1963}
G.~Dantzig.
\newblock {\em {Linear Programming and Extensions}}.
\newblock Princeton University Press, 1963.

\bibitem{Ehlers2017}
R.~Ehlers.
\newblock {Formal Verification of Piece-Wise Linear Feed-Forward Neural
  Networks}.
\newblock In {\em Proc. 15th Int. Symp. on Automated Technology for
  Verification and Analysis (ATVA)}, pages 269--286, 2017.

\bibitem{Eh17}
R.~Ehlers.
\newblock {Formal Verification of Piece-Wise Linear Feed-Forward Neural
  Networks}.
\newblock In {\em Proc. 15th Int. Symp. on Automated Technology for
  Verification and Analysis (ATVA)}, pages 269--286, 2017.

\bibitem{ElGoKa20}
Y.~Elboher, J.~Gottschlich, and G.~Katz.
\newblock {An Abstraction-Based Framework for Neural Network Verification}.
\newblock In {\em Proc. 32nd Int. Conf. on Computer Aided Verification (CAV)},
  pages 43--65, 2020.

\bibitem{ElKaKaSc21}
T.~Eliyahu, Y.~Kazak, G.~Katz, and M.~Schapira.
\newblock {Verifying Deep-RL-Driven Systems}.
\newblock In {\em Proc. Annual Conf. of the ACM Special Interest Group on Data
  Communication on the Applications, Technologies, Architectures, and Protocols
  for Computer Communication (SIGCOMM)}, 2021.

\bibitem{GeMiDrTsChVe18}
T.~Gehr, M.~Mirman, D.~Drachsler-Cohen, E.~Tsankov, S.~Chaudhuri, and
  M.~Vechev.
\newblock {AI2: Safety and Robustness Certification of Neural Networks with
  Abstract Interpretation}.
\newblock In {\em Proc. 39th IEEE Symposium on Security and Privacy (S\&P)},
  2018.

\bibitem{GoFeMaBaKa20}
S.~Gokulanathan, A.~Feldsher, A.~Malca, C.~Barrett, and G.~Katz.
\newblock {Simplifying Neural Networks using Formal Verification}.
\newblock In {\em Proc. 12th NASA Formal Methods Symposium (NFM)}, pages
  85--93, 2020.

\bibitem{GoAdKeKa20}
B.~Goldberger, Y.~Adi, J.~Keshet, and G.~Katz.
\newblock {Minimal Modifications of Deep Neural Networks using Verification}.
\newblock In {\em Proc. 23rd Int. Conf. on Logic for Programming, Artificial
  Intelligence and Reasoning (LPAR)}, pages 260--278, 2020.

\bibitem{GoBeCo16}
I.~Goodfellow, Y.~Bengio, and A.~Courville.
\newblock {\em {Deep Learning}}.
\newblock MIT Press, 2016.

\bibitem{GoKaPaBa18}
D.~Gopinath, G.~Katz, C.~P\v{a}s\v{a}reanu, and C.~Barrett.
\newblock {DeepSafe: A Data-driven Approach for Assessing Robustness of Neural
  Networks}.
\newblock In {\em Proc. 16th. Int. Symposium on on Automated Technology for
  Verification and Analysis (ATVA)}, pages 3--19, 2018.

\bibitem{HaMaDa15}
S.~Han, H.~Mao, and W.~Dally.
\newblock {Deep Compression: Compressing Deep Neural Networks with Pruning,
  Trained Quantization and Huffman Coding}, 2015.
\newblock Technical Report. \url{http://arxiv.org/abs/1510.00149}.

\bibitem{HuKwWaWu17}
X.~Huang, M.~Kwiatkowska, S.~Wang, and M.~Wu.
\newblock {Safety Verification of Deep Neural Networks}.
\newblock In {\em Proc. 29th Int. Conf. on Computer Aided Verification (CAV)},
  pages 3--29, 2017.

\bibitem{HuCoSoElBe16}
I.~Hubara, M.~Courbariaux, D.~Soudry, R.~El-Yaniv, and Y.~Bengio.
\newblock {Binarized Neural Networks}.
\newblock In {\em Proc. 30th Conf. on Neural Information Processing Systems
  (NIPS)}, pages 4107--4115, 2016.

\bibitem{HuCoSoElBe17}
I.~Hubara, M.~Courbariaux, D.~Soudry, R.~El-Yaniv, and Y.~Bengio.
\newblock {Quantized Neural Networks: Training Neural Networks with Low
  Precision Weights and Activations}.
\newblock {\em Journal of Machine Learning Research}, 18(1):6869--6898, 2017.

\bibitem{IaHaMoAsDaKe16}
F.~Iandola, S.~Han, M.~Moskewicz, K.~Ashraf, W.~Dally, and K.~Keutzer.
\newblock {SqueezeNet: AlexNet-level Accuracy with 50x Fewer Parameters and $<$
  0.5MB Model Size}, 2016.
\newblock Technical Report. \url{http://arxiv.org/abs/1602.07360}.

\bibitem{JaBaKa20}
Y.~Jacoby, C.~Barrett, and G.~Katz.
\newblock {Verifying Recurrent Neural Networks using Invariant Inference}.
\newblock In {\em Proc. 18th Int. Symposium on Automated Technology for
  Verification and Analysis (ATVA)}, pages 57--74, 2020.

\bibitem{JuLoBrOwKo16}
K.~Julian, J.~Lopez, J.~Brush, M.~Owen, and M.~Kochenderfer.
\newblock {Policy Compression for Aircraft Collision Avoidance Systems}.
\newblock In {\em Proc. 35th Digital Avionics Systems Conf. (DASC)}, pages
  1--10, 2016.

\bibitem{KaBaDiJuKo17}
G.~Katz, C.~Barrett, D.~Dill, K.~Julian, and M.~Kochenderfer.
\newblock {Reluplex: An Efficient SMT Solver for Verifying Deep Neural
  Networks}.
\newblock In {\em Proc. 29th Int. Conf. on Computer Aided Verification (CAV)},
  pages 97--117, 2017.

\bibitem{KaBaDiJuKo17Fvav}
G.~Katz, C.~Barrett, D.~Dill, K.~Julian, and M.~Kochenderfer.
\newblock {Towards Proving the Adversarial Robustness of Deep Neural Networks}.
\newblock In {\em Proc. 1st Workshop on Formal Verification of Autonomous
  Vehicles (FVAV)}, pages 19--26, 2017.

\bibitem{KaBaDiJuKo21}
G.~Katz, C.~Barrett, D.~Dill, K.~Julian, and M.~Kochenderfer.
\newblock {Reluplex: a Calculus for Reasoning about Deep Neural Networks}.
\newblock {\em Formal Methods in System Design (FMSD)}, 2021.
\newblock To appear.

\bibitem{Marabou2019}
G.~Katz, D.~Huang, D.~Ibeling, K.~Julian, C.~Lazarus, R.~Lim, P.~Shah,
  S.~Thakoor, H.~Wu, A.~Zelji\'c, D.~Dill, M.~Kochenderfer, and C.~Barrett.
\newblock {The Marabou Framework for Verification and Analysis of Deep Neural
  Networks}.
\newblock In {\em Proc. 31st Int. Conf. on Computer Aided Verification (CAV)},
  pages 443--452, 2019.

\bibitem{KaBaKaSc19}
Y.~Kazak, C.~Barrett, G.~Katz, and M.~Schapira.
\newblock {Verifying Deep-RL-Driven Systems}.
\newblock In {\em Proc. 1st ACM SIGCOMM Workshop on Network Meets AI \& ML
  (NetAI)}, pages 83--89, 2019.

\bibitem{KoLoJaBl20}
B.~K\"{o}nighofer, F.~Lorber, N.~Jansen, and R.~Bloem.
\newblock {Shield Synthesis for Reinforcement Learning}.
\newblock In {\em Proc. Int. Symposium On Leveraging Applications of Formal
  Methods, Verification and Validation (ISoLA)}, pages 290--306, 2020.

\bibitem{KuKaGoJuBaKo18}
L.~Kuper, G.~Katz, J.~Gottschlich, K.~Julian, C.~Barrett, and M.~Kochenderfer.
\newblock {Toward Scalable Verification for Safety-Critical Deep Networks},
  2018.
\newblock Technical Report. \url{https://arxiv.org/abs/1801.05950}.

\bibitem{ourCode}
O.~Lahav and G.~Katz.
\newblock {Code: Pruning and Slicing Neural Networks using Formal
  Verification}, 2021.
\newblock \url{https://github.com/vbcrlf/redy}.

\bibitem{liebenwein2021lost}
L.~Liebenwein, C.~Baykal, B.~Carter, D.~Gifford, and D.~Rus.
\newblock {Lost in Pruning: The Effects of Pruning Neural Networks beyond Test
  Accuracy}, 2021.
\newblock Technical Report. \url{https://arxiv.org/abs/2103.03014}.

\bibitem{LiArLaBaKo19}
C.~Liu, T.~Arnon, C.~Lazarus, C.~Barrett, and M.~Kochenderfer.
\newblock {Algorithms for Verifying Deep Neural Networks}, 2019.
\newblock Technical Report. \url{http://arxiv.org/abs/1903.06758}.

\bibitem{LoMa17}
A.~Lomuscio and L.~Maganti.
\newblock {An Approach to Reachability Analysis for Feed-Forward ReLU Neural
  Networks}, 2017.
\newblock Technical Report. \url{http://arxiv.org/abs/1706.07351}.

\bibitem{Gurobi}
G.~Optimization.
\newblock {The Gurobi MILP Solver}, 2021.
\newblock \url{https://www.gurobi.com/}.

\bibitem{RaOrReFa16}
M.~Rastegari, V.~Ordonez, J.~Redmon, and A.~Farhadi.
\newblock {XNOR-Net: Imagenet Classification using Binary Convolutional Neural
  Networks}.
\newblock In {\em Proc. 14th European Conf. on Computer Vision (ECCV)}, pages
  525--542, 2016.

\bibitem{SiHuMaGuSiVaScAnPaLaDi16}
D.~Silver, A.~Huang, C.~Maddison, A.~Guez, L.~Sifre, G.~Van Den~Driessche,
  J.~Schrittwieser, I.~Antonoglou, V.~Panneershelvam, M.~Lanctot, and
  S.~Dieleman.
\newblock {Mastering the Game of Go with Deep Neural Networks and Tree Search}.
\newblock {\em Nature}, 529(7587):484--489, 2016.

\bibitem{SiZi14}
K.~Simonyan and A.~Zisserman.
\newblock {Very Deep Convolutional Networks for Large-Scale Image Recognition},
  2014.
\newblock Technical Report. \url{http://arxiv.org/abs/1409.1556}.

\bibitem{StWuZeJuKaBaKo20}
C.~Strong, H.~Wu, A.~Zelji\'c, K.~Julian, G.~Katz, C.~Barrett, and
  M.~Kochenderfer.
\newblock {Global Optimization of Objective Functions Represented by ReLU
  networks}, 2020.
\newblock Technical Report. \url{http://arxiv.org/abs/2010.03258}.

\bibitem{SzZaSuBrErGoFe13}
C.~Szegedy, W.~Zaremba, I.~Sutskever, J.~Bruna, D.~Erhan, I.~Goodfellow, and
  R.~Fergus.
\newblock {Intriguing Properties of Neural Networks}, 2013.
\newblock Technical Report. \url{http://arxiv.org/abs/1312.6199}.

\bibitem{TjXiTe17}
V.~Tjeng, K.~Xiao, and R.~Tedrake.
\newblock {Evaluating Robustness of Neural Networks with Mixed Integer
  Programming}, 2017.
\newblock Technical Report. \url{http://arxiv.org/abs/1711.07356}.

\bibitem{TrBkJo20}
H.~Tran, S.~Bak, and T.~Johnson.
\newblock {Verification of Deep Convolutional Neural Networks Using
  ImageStars}.
\newblock In {\em Proc. 32nd Int. Conf. on Computer Aided Verification (CAV)},
  pages 18--42, 2020.

\bibitem{WaPeWhYaJa18}
S.~Wang, K.~Pei, J.~Whitehouse, J.~Yang, and S.~Jana.
\newblock {Formal Security Analysis of Neural Networks using Symbolic
  Intervals}.
\newblock In {\em Proc. 27th USENIX Security Symposium}, 2018.

\bibitem{WeZhChSoHsBoDhDa18}
T.-W. Weng, H.~Zhang, H.~Chen, Z.~Song, C.-J. Hsieh, D.~Boning, I.~Dhillon, and
  L.~Daniel.
\newblock {Towards Fast Computation of Certified Robustness for ReLU Networks},
  2018.
\newblock Technical Report. \url{http://arxiv.org/abs/1804.09699}.

\bibitem{WuOzZeIrJuGoFoKaPaBa20}
H.~Wu, A.~Ozdemir, A.~Zelji\'c, A.~Irfan, K.~Julian, D.~Gopinath, S.~Fouladi,
  G.~Katz, C.~P\u{a}s\u{a}reanu, and C.~Barrett.
\newblock {Parallelization Techniques for Verifying Neural Networks}.
\newblock In {\em Proc. 20th Int. Conf. on Formal Methods in Computer-Aided
  Design (FMCAD)}, pages 128--137, 2020.

\bibitem{ZhShGuGuLeNa20}
H.~Zhang, M.~Shinn, A.~Gupta, A.~Gurfinkel, N.~Le, and N.~Narodytska.
\newblock {Verification of Recurrent Neural Networks for Cognitive Tasks via
  Reachability Analysis}.
\newblock In {\em Proc. 24th Conf. of European Conference on Artificial
  Intelligence (ECAI)}, pages 1690--1697, 2020.

\end{thebibliography}
}

\newpage
\onecolumn

\begin{appendices}

\section{$\errorMinimizerFunction{}(x)$ Function Formula Proof}
\label{appendix:relax_analysis}

\subsection{Introduction}

In Section \ref{sec:linear_funcs} we introduced $\errorMinimizerFunction{}(x)$:
\[
  \errorMinimizerFunction{}(x) = \frac{u}{u - l} \cdot x
  + \frac{-l\cdot u}{2(u - l)}.
\]

\subsection{Proof}

\begin{figure}[ht]
\begin{center}
\includegraphics[scale=0.7]{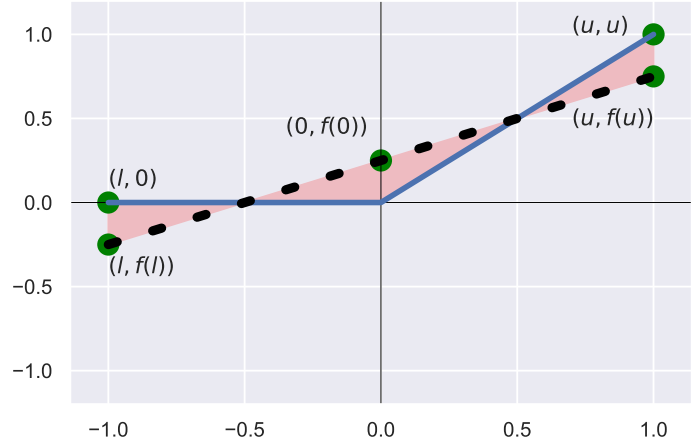}
\end{center}
\caption {$\errorMinimizerFunction{}(x)$ function illustration with a few relevant points. \label{fig:relaxed_appendix}}
\end{figure}

We would like to find $f(x) = ax+b$ where the maximum error is minimized.
The first observation is that in order $f(x)$ to minimize the maximum error,
it must be below $(l, 0)$ and $(u, u)$, but above $(0, f(0))$ (otherwise,
the maximum error could be trivially improved). See Figure \ref{fig:relaxed_appendix}.

Follows from this observation, that our goal is to minimize the following term:
\[
	max(\{|0 - f(l)|, |0 - f(0)|, |u - f(u)|\})
\]

Re-writing the term with $f(x)$ definition we get:
\[
	max(\{-al-b, b, (1-a)u-b\})
\]

Observe that $a = \frac{u}{u - l}$, $b = \frac{-l\cdot u}{2(u - l)}$ is a minimum: any $\pm\epsilon$ change to $a$ or $b$ will result in the increasement of one of the three terms.

In addition, note that in this case the 3 terms inside the $max$ are equal and thus the maximum error is $b$.

\newpage
\onecolumn

\section{Simultanous Neuron Removal Error Bounds}
\label{appendix:sim_error_bounds}

\subsection{Introduction}

Terminology:

\[forward_i^{(0)} = \mbox{input number \#i}\]
\[backward_j^{(i)} = \mbox{hidden layer j neuron \#i backward value}\]
\[= bias_j^{(i)} + \sum_k w_{j,k}^{(i)} \cdot forward^{(i-1)}_{(k)}\]
\[forward_j^{(i)} = func_j^{(i)}(backward_j^{(j)} )\]
$\mbox{where} func_j^{(i)} = \mbox{ReLU or Identity or Zero}$. Last layer has func = Identity, and every neuron's ReLU can be potentially replaced with Identity or Zero in the amended network. If the variables have \(\overline{overline}\), it means they are variables of the amended network. Assuming a subset (or all) of the neurons are redundant in the input space, e.g., their $func_j^{(i)}$ may have been replaced with one of Identity, Zero or positively-sloped $f(x)$ and still not be far from the original output.

Specifically: If the neuron is to be replaced with Zero we require:
\[ backward_j^{(i)} \leq \epsilon_j^{(i)} \]
If is to be replaced with Identity we require:
\[ -\epsilon_j^{(i)} \leq backward_j^{(i)} \]
And if is to be replaced with any positively-sloped linear $f(x)$ (for example, $\errorMinimizerFunction{}(x)$) we require the maximum error to be $\epsilon_j^{(i)}$ at maximum (in the case of $\errorMinimizerFunction{}(x)$ we will have $\epsilon_j^{(i)} = \frac{-l\cdot u}{2(u - l)}$).
This leads to an error of no more than $\epsilon_j^{(i)}$ on the forward value of this neuron.

We would like to bound the error in the network when we replace all these neurons' $func$s simultaneously.

\subsection{Input layer has no errors}
\noindent
The input layer has no error, and so:
\[ forward_j^{(0)} - 0 \leq \overline{forward}_j^{(0)} \leq forward_j^{(0)} + 0 \]
So we set the lower and upper error bounds of the input layer to zero --- ${\bf err}_j^{(l,0)} = 0$ and ${\bf err}_j^{(u,0)} = 0$.

\subsection{Bounding hidden layer's backward value}
\noindent
Each neuron backward value holds:
\[ \overline{backward}_j^{(i)} = bias_j^{(i)} + \sum_k w_{j,k}^{(i)} \cdot \overline{forward}^{(i-1)}_{k} = \]
\[ bias_j^{(i)} + \sum_{k,w \geq 0} w_{j,k}^{(i)} \cdot \overline{forward}^{(i-1)}_{k} + \sum_{k,w \leq 0} w_{j,k}^{(i)} \cdot \overline{forward}^{(i-1)}_{k} = (*) \]

\noindent
Upper bound for $(*)$:
\[ (*) \leq bias_j^{(i)} + \sum_{k,w \geq 0} w_{j,k}^{(i)} \cdot (forward^{(i-1)}_{j} + {\bf err}_k^{(u,i-1)}) + \sum_{k,w \leq 0} w_{j,k}^{(i)} \cdot (forward^{(i-1)}_{j} - {\bf err}_k^{(l,i-1)}) \]
\[ = bias_j^{(i)} + \sum_k w_{j,k}^{(i)} \cdot forward^{(i-1)}_{j} + \sum_{k,w \geq 0} w_{j,k}^{(i)} \cdot {\bf err}_k^{(u,i-1)} - \sum_{k,w \leq 0} w_{j,k}^{(i)} \cdot {\bf err}_k^{(l,i-1)} \]
\[ = bias_j^{(i)} + \sum_k w_{j,k}^{(i)} \cdot forward^{(i-1)}_{j} + \sum_{k,w \geq 0} w_{j,k}^{(i)} \cdot {\bf err}_k^{(u,i-1)} - \sum_{k,w \leq 0} w_{j,k}^{(i)} \cdot {\bf err}_k^{(l,i-1)} \]
\[ = backward^{(i)}_{j} + \sum_{k,w \geq 0} w_{j,k}^{(i)} \cdot {\bf err}_k^{(u,i-1)} - \sum_{k,w \leq 0} w_{j,k}^{(i)} \cdot {\bf err}_k^{(l,i-1)} \]
\[ = backward^{(i)}_{j} + B^{(i)}_{j} \]

\noindent
Lower bound for $(*)$:
\[ (*) \geq backward^{(i)}_{j} - ( \sum_{k,w \geq 0} w_{j,k}^{(i)} \cdot {\bf err}_k^{(l,i-1)} - \sum_{k,w \leq 0} w_{j,k}^{(i)} \cdot {\bf err}_k^{(u,i-1)} ) = backward^{(i)}_{j} - A^{(i)}_{j} \]

\noindent
Overall we have:
\[ backward^{(i)}_{j} - A^{(i)}_{j} \leq \overline{backward}_j^{(i)} \leq backward^{(i)}_{j} + B^{(i)}_{j} \]

\subsection{Bounding hidden layer's forward value}
\noindent
Split into cases:

\noindent
{\bf 1. Neuron is left unchanged}, e.g. $\overline{func}_j^{(i)} = ReLU$
\[
    \overline{forward}_j^{(i)} = ReLU(\overline{backward}_j^{(j)}) \geq ReLU({backward}_j^{(i)} - A_j^{(i)}) \geq ReLU({backward}_j^{(i)}) - A_j^{(i)}
\]
\[
    = {forward}_j^{(i)} - A_j^{(i)}
\]
\noindent
The last inequality is true because $-A_j^{(i)} \leq 0$. Upper bound:
\[
    \overline{forward}_j^{(i)} = ReLU(\overline{backward}_j^{(j)}) \leq ReLU({backward}_j^{(i)} + B_j^{(i)}) \leq ReLU({backward}_j^{(i)}) + B_j^{(i)}
\]
\[
    = {forward}_j^{(i)} + B_j^{(i)}
\]
\noindent
The last inequality is true because $B_j^{(i)} \geq 0$.

\noindent
So we set:
\[{\bf err}_j^{(l,i)} = A_j^{(i)} \mbox{,\ \ } {\bf err}_j^{(u,i)} = B_j^{(i)} \]

\noindent
{\bf 2. Neuron is replaced with Identity}, e.g. $\overline{func}_j^{(i)} = Identity$

\noindent
We have (from the assumptions):
\[ -\epsilon_j^{(i)} \leq backward_j^{(i)} \]
And we also have the backward value bound:
\[ backward^{(i)}_{j} - A^{(i)}_{j} \leq \overline{backward}_j^{(i)} \leq backward^{(i)}_{j} + B^{(i)}_{j} \]
Combining these:
\[ -\epsilon_j^{(i)} - A^{(i)}_{j} \leq \overline{backward}_j^{(i)} = \overline{forward}_j^{(i)} \leq backward^{(i)}_{j} + B^{(i)}_{j} \]

\noindent
If $backward_j^{(i)} \leq 0$ then we have $forward_j^{(i)} = 0$ and so:
\[ forward_j^{(i)} - (A^{(i)}_{j} + \epsilon_j^{(i)}) \leq \overline{forward}_j^{(i)} \leq backward^{(i)}_{j} + B^{(i)}_{j} \leq forward_j^{(i)} + B^{(i)}_{j} \]
And if $backward_j^{(i)} \geq 0$ then we have $forward_j^{(i)} = backward_j^{(i)}$ and so:
\[ forward^{(i)}_{j} - A^{(i)}_{j} \leq \overline{forward}_j^{(i)} \leq forward^{(i)}_{j} + B^{(i)}_{j} \]
So overall we set:
\[{\bf err}_j^{(l,i)} = A^{(i)}_{j} + \epsilon_j^{(i)} \mbox{,\ \ } {\bf err}_j^{(u,i)} = B_j^{(i)} \]

\noindent
{\bf 3. Neuron is replaced with Zero}, e.g. $\overline{func}_j^{(i)} = Zero$

\noindent
We have (from the assumptions):
\[ backward_j^{(i)} \leq \epsilon_j^{(i)} \]
\[ \Downarrow \]
\[ forward_j^{(i)} \leq \epsilon_j^{(i)} \]
\[ \Downarrow \]
\[ forward_j^{(i)} - \epsilon_j^{(i)} \leq 0 = \overline{forward}_j^{(i)} \leq forward_j^{(i)} + 0 \]

\noindent
So overall we set:
\[{\bf err}_j^{(l,i)} = \epsilon_j^{(i)} \mbox{,\ \ } {\bf err}_j^{(u,i)} = 0 \]

\noindent
{\bf 4. Neuron is replaced with positively sloped $f(x)$}, e.g. $\overline{func}_j^{(i)} = f(x)$

\noindent
We have:
\[ \overline{forward}_j^{(i)} = f(\overline{backward}_j^{(i)}) \]
Combining this with the backward value bound we get (using the fact $f(x)$ is positively sloped):
\[ f(backward^{(i)}_{j} - A^{(i)}_{j}) \leq \overline{forward}_j^{(i)} \leq f(backward^{(i)}_{j} + B^{(i)}_{j}) \]
Using $f(x)$ linearity:
\[ -\epsilon_j^{(i)} - f(A^{(i)}_{j}) \leq f(backward^{(i)}_{j}) - f(A^{(i)}_{j}) \leq \overline{forward}_j^{(i)} \leq f(backward^{(i)}_{j}) + f(B^{(i)}_{j}) \leq \epsilon_j^{(i)} + f(B^{(i)}_{j}) \]
So we set:
\[{\bf err}_j^{(l,i)} = f(A^{(i)}_{j}) + \epsilon_j^{(i)} \mbox{,\ \ } {\bf err}_j^{(u,i)} = f(B_j^{(i)}) + \epsilon_j^{(i)} \]

\subsection{Output Layer}
\noindent
We have:
\[ backward^{(i)}_{j} - A^{(i)}_{j} \leq \overline{backward}_j^{(i)} \leq backward^{(i)}_{j} + B^{(i)}_{j} \]
And:
\[ forward^{(i)}_{j} = backward^{(i)}_{j} \mbox{ , } \overline{forward}^{(i)}_{j} = \overline{backward}^{(i)}_{j} \]
So we set:
\[{\bf err}_j^{(l,i)} = A_j^{(i)} \mbox{,\ \ } {\bf err}_j^{(u,i)} = B_j^{(i)} \]

\end{appendices}

\end{document}